\documentclass{article}

\usepackage[T1]{fontenc}
\usepackage[utf8]{inputenc}

\usepackage{microtype}
\usepackage{graphicx}
\usepackage{booktabs}      
\usepackage{float}

\usepackage{listings}
\lstdefinestyle{promptstyle}{%
  basicstyle=\ttfamily\footnotesize,%
  breaklines=true,%
  breakautoindent=false,%
  columns=fullflexible,%
  keepspaces=true,%
  showstringspaces=false,%
  frame=single,%
  xleftmargin=1em,%
  extendedchars=true,%
  literate=%
    {ä}{{\"a}}1 {ö}{{\"o}}1 {ü}{{\"u}}1 {Ä}{{\"A}}1 {Ö}{{\"O}}1 {Ü}{{\"U}}1%
    {ß}{{\ss}}1 {§}{{\S}}1%
    {„}{{,,}}1 {“}{{``}}1 {”}{{''}}1 {–}{{--}}1 {—}{{---}}1 {…}{{\ldots}}1%
}

\usepackage{hyperref}

\usepackage[accepted]{icml2026}
\makeatletter 
\renewcommand{\Notice@String}{Accepted at the ICML 2026 Workshop on AI for Law, Seoul, South Korea.} 
\makeatother
\usepackage{amsmath}
\usepackage{amssymb}
\usepackage{mathtools}
\usepackage{amsthm}

\usepackage[capitalize,noabbrev]{cleveref}

\theoremstyle{definition}

\newtheorem{example}{Example}          
\theoremstyle{remark}

\usepackage[disable]{todonotes}

\icmltitlerunning{Classifying Interpretive Canons in German Constitutional Court Decisions}

\begin{document}

\twocolumn[
\icmltitle{Classifying Interpretive Canons at the Sentence Level: \\ A Benchmark from the German Federal Constitutional Court}

\begin{icmlauthorlist}
\icmlauthor{Felix Ringe}{fub}
\end{icmlauthorlist}

\icmlaffiliation{fub}{Department of Law, Freie Universität Berlin, Berlin, Germany}

\icmlcorrespondingauthor{Felix Ringe}{felix.ringe@fu-berlin.de}

\icmlkeywords{legal NLP, argument mining, LLM evaluation, benchmark, statutory interpretation}

\vskip 0.3in
]

\printAffiliationsAndNotice{}  

\begin{abstract}
Judicial reasoning remains challenging for large language models (LLMs) to analyze. This paper contributes a sentence-level benchmark for evaluating the ability of LLMs to classify interpretive canons as articulated by Larenz in the tradition of Savigny. Our contributions are threefold. First, we operationalize this conception of interpretation as classification criteria. Second, we provide a dataset of decisions of the German Federal Constitutional Court annotated at the sentence level. Third, we report baseline evaluations of four LLMs from three model families under expert hand-written prompts, compared against prompts optimized with Genetic-Pareto (GEPA). Mean $F_1$ over the seven binary subtasks clusters between 70.4 and 79.2 across models, with grammatical interpretation usually the easiest canon to identify and systematic interpretation usually the hardest; under the tested configuration, GEPA-optimized prompts do not systematically outperform the hand-written ones, suggesting that the expert prompts provide a meaningful baseline.
\end{abstract}

\section{Introduction}
\label{sec:intro}

The systematic identification of the types of arguments that courts use to justify their decisions has long been a focus of empirical legal scholarship. These efforts are commonly motivated by the idea that the core of just judicial decision-making lies in upholding the guarantee of equal treatment. This guarantee, in turn, finds expression in a transparent justification of decisions, and in particular in the consistent use of a recognized set of interpretive arguments \citep{Raisch1988}. Such consistency is also said to make judicial decision-making more predictable and to constrain judicial discretion \citep{GorsuchNeilM2016Olab}.

The capacity to study these arguments at scale has long been constrained by the manual effort required, forcing a careful selection of the materials to be analyzed \citep{Mendelson2018}. Reliable classification of these argument types by LLMs promises to dramatically decrease the cost of such analysis and open up a host of substantive research questions. For instance, this prospect has motivated work testing long-standing narratives about the historical degree of formalism in courts for the United States \citep{Stiglitz2024} and the Czech Republic \citep{Koref2026} as well as analyses of reasoning styles of individual justices and the relationship between legal reasoning and other judicial features \citep{Thalken2026}.

However, prior work has taken the paragraph as its modeling unit, which identifies argument types within paragraphs but not their precise locations. This paper develops a finer-grained approach to classifying interpretation, operating at the sentence level. The analysis is built on the four interpretive canons of grammatical, systematic, historical, and objective-teleological interpretation, usually traced to Friedrich Carl von Savigny and given their modern form, most prominently, by Karl Larenz \citep{Larenz1991}. This conception has found resonance in many jurisdictions through frequent translation of his work, with editions published in Spain \citep{Larenz2023}, Portugal \citep{Larenz2014}, the People's Republic of China \citep{Larenz2020}, and South Korea \citep{Larenz2026}, among others.

This paper makes three contributions: (1) it operationalizes Larenz's conception of interpretation into granular classification criteria (\cref{sec:task}), (2) it contributes an expert-annotated dataset of decisions of the German Federal Constitutional Court that follows these criteria and records individual reasons for edge cases (\cref{sec:dataset}), and (3) it reports baseline evaluations of four LLMs from three model families on the resulting subtasks (\cref{sec:experiments,sec:results}).\footnote{\label{fn:wandb}Code available on \href{https://github.com/KensingtonOscupant/classifying-interpretive-canons}{GitHub}, datasets on \href{https://huggingface.co/collections/felix453/interpretive-canons}{Hugging Face}, and evaluation runs on \href{https://wandb.ai/icml-2026-ai4law/classifying-interpretive-canons-camera-ready/}{Weights \& Biases}. Agent trajectories for successful reproductions of the \href{https://hub.harborframework.com/jobs/35917940-c695-43ab-bdd9-17ddea56db05/trials/e8b5e216-2f0c-4cc3-b40f-d291e64c1833}{scores} and \href{https://hub.harborframework.com/jobs/676116e1-8968-4245-a097-c37e6c845ed6/trials/b0f8c8ef-d598-49c1-a5dd-a4acb8d364e9}{datasets} can be inspected on Harbor Hub. For more details on the datasets, see \cref{sec:dataset}.} 

We find that an AI agent can reproduce both the reported scores and benchmark datasets exactly from the respective input data and our methods descriptions, without access to our code (\cref{sec:reproduction}).

\section{Task Methodology}
\label{sec:task}

To examine exactly where interpretation occurs in decisions, it is necessary to establish a working definition that allows for classification to be carried out. Given the extensive literature on the concept of interpretation, one might expect formulating such a definition to be straightforward. It is not, as there is no standardized definition: \citet{Lueders2025} report the same difficulty for defining proportionality, a related concept central to German constitutional reasoning. Insightful general remarks on the concept of interpretation can, however, be found in Karl Larenz's \emph{Methodenlehre der Rechtswissenschaft}:
\begin{quote}
``To `interpret' a text thus means to decide in favor of one among
several possible readings on the basis of considerations
[\ldots]``~\citep[p.~204]{Larenz1991}\footnote{In the original: ``Einen Text `auslegen', hei\ss{}t also, sich
f\"ur eine unter mehreren m\"oglichen Deutungen aufgrund von
\"Uberlegungen zu entscheiden [\ldots]''. Translation ours.}
\end{quote}

Several prerequisites can be derived from this definition. First, one must decide on a reading. Second, this must involve certain considerations. Third, the decision in favor of the reading must follow precisely from those considerations. In statutory interpretation, these considerations regularly draw on the interpretive canons which this work examines. All of the above requirements will be examined more closely in what follows.

\subsection{Reading}
\label{sec:reading}

A reading within the meaning of this study is an assertion that the court itself advances and whose object is the claim that a particular statutory provision, in the abstract, has a particular content. A detailed explanation follows in the sections below. For the study of interpretive canons, the precise determination of the reading is decisive, because readings are the reference points of the interpretive canons. Which of the interpretive canons are present can depend on the reading from which the consideration is viewed:

\begin{example}\label{ex:redetermination}
``Section 46(1), first sentence, no.~2 BWahlG provides for the loss of the parliamentary seat in the case of a `redetermination' of the election result.''\footnote{Translations of excerpts of court decisions in this paper are ours throughout.} (BVerfGE 124, 1 (15))\footnote{Decisions are cited to the court's official reports, \emph{Entscheidungen des Bundesverfassungsgerichts} (BVerfGE), by volume, the page on which the decision begins, and, in parentheses, the specific page referred to; e.g., BVerfGE 124, 1 (15) is volume 124, decision beginning at page~1, cited at page~15.}
\end{example}

In \cref{ex:redetermination}, one might assume at first glance that, because of the word ``redetermination'' in quotation marks, this is a grammatical interpretation of Section~46 BWahlG (\emph{Bundeswahlgesetz}, the Federal Electoral Act). This picture changes once the preceding sentence is brought into view. The expanded passage reads as follows:

\begin{example}\label{ex:electoral-act}
``There are, however, indications supporting an interpretation to the effect that the `electoral act' within the meaning of Section~37 BWahlG is completed on the evening of the main election and that a new `electoral act' begins with the supplementary election: Section~46(1), first sentence, no.~2 BWahlG provides for the loss of the parliamentary seat in the case of a `redetermination' of the election result.'' (BVerfGE 124, 1 (15)) \end{example}

In \cref{ex:electoral-act}, the reference point is no longer Section~46 BWahlG, as in \cref{ex:redetermination}, but Section~37 BWahlG. From the perspective of Section~37 BWahlG, the statement about Section~46 BWahlG constitutes systematic interpretation. From this vantage point, then, the statement about Section~46 BWahlG shifts from a grammatical interpretation to a systematic one.

The above definition of a reading identifies four prerequisites, which will be discussed in more detail below.

\subsubsection{Abstract}

First, the assertion must be abstract. A statement is abstract if it claims validity not only for the unique set of facts underlying the decision, but for an indeterminate range of situations in which the statutory provision or legal principle applies. Statutory interpretation is always concerned with determining the meaning of a norm independently of the concrete facts. This criterion also marks the distinction from subsumption, which brings a set of facts under a norm. The boundary is at times not easy to draw.

\begin{example}\label{ex:stpo-letter}
``Where the letter of a person held in pre-trial detention is, as here, seized pursuant to the analogous application of Section~108 StPO and forwarded to the public prosecutor's office for further action, the person concerned may apply to the public prosecutor's office for the return of the letter, may at any time request a decision by the competent judge (Section~98(2) StPO), and may lodge a complaint against any seizure during the preliminary proceedings (Section~304(1) StPO).''
(BVerfGE 57, 170 (181))
\end{example}

In \cref{ex:stpo-letter}, the phrase ``as here'' establishes a concrete connection to the facts, which weighs against an abstract statement. The rest of the sentence, however, states a conclusion about the meaning of Section~98(2) and Section~304(1) StPO (\emph{Strafprozessordnung}, the Code of Criminal Procedure) that applies to other cases as well. Since the connection to the facts is established only incidentally, and the sentence does not aim to make the statement solely for this one situation, it is more apt to read the statement as abstract.

\subsubsection{Assertion}
Furthermore, the statement must be an assertion. An assertion is a statement against which a meaningful counterposition can be formed. This is not the case where the statement is purely descriptive, for instance the verbatim reproduction of the statutory text. This rests on the consideration that a reading fills the statutory provision precisely with content, and that this requires an assertion going beyond the mere statement of the wording. While an argument employing an interpretive canon, for example in the context of literal interpretation, may well fall back on merely repeating the wording, this does not suffice for a reading.\footnote{Of the four
prerequisites, assertion is the one we do not operationalize as a stand-alone classification subtask in this work; see \cref{sec:dataset}.}

\subsubsection{Reference Point: A Specific Statutory Provision with Determinate Content}
\label{sec:reference-point}

Furthermore, the assertion must relate to a specific statutory provision. This restriction is helpful because it fixes the object of interpretation on an unambiguous reference point. In particular, it excludes from consideration those statements that at first appear interpretive because they invoke a legal principle or general principle, but which in fact have no interpretable reference point in the written law.

Naming a provision is not sufficient, however: the assertion must also determine that provision's content. It does so when it has the effect that a range of cases falls under the provision while others do not. An assertion that aims instead at declaring the provision as a whole void, valid, or inapplicable does not state a determinate content. The practical purpose of this second half of the criterion is to exclude sentences whose object is merely the compatibility of one norm with another, for example the steps of a proportionality or constitutionality assessment, since such statements concern the relation between two norms rather than the content of either one.

\begin{example}\label{ex:compatibility}
``Section~901 ZPO is consequently inapplicable where the debtor's inability to pay is established, and to that extent cannot violate Art.~2(2), second sentence, of the Basic Law.'' (BVerfGE 61, 126)
\end{example}

\cref{ex:compatibility} names two provisions precisely, yet it fixes the content of neither. It states how the one relates to the other, which leaves open which cases Section~901 ZPO (\emph{Zivilprozessordnung}, the Code of Civil Procedure) covers. We treat reference point and determinate content as a single criterion because they share a reference point and are decided together in one pass over the sentence; the classification subtask in \cref{sec:dataset} accordingly asks for the provision the content of which the sentence determines, and returns nothing when no such provision exists.

\subsubsection{Self-advanced}
An assertion is self-advanced when it has both its starting point and its end point in the reasoning of the deciding panel. This is not the case when the court bases an assertion on a precedent. That precedents, too, might fall under the concept of interpretation does not seem entirely far-fetched, since they often themselves contain passages in which interpretation is carried out. By referring to such a passage, the interpretation undertaken there is implicitly perpetuated, at least insofar as the court invokes the precedent approvingly. However, it does not appear certain that the citation of a precedent permits an inference that the panel approves of every interpretation undertaken in that precedent. Where an assertion is supported by several sources, for example a scholarly source, a legislative explanatory memorandum, and a precedent, not all of which are precedents, the assertion is assessed, in the panel's favor, as self-advanced.

\subsection{Considerations}
\label{sec:considerations}

Interpretation means choosing one of several possible readings on the basis of considerations. This gives a first criterion: a reading must be reached ``on the basis of considerations'', i.e., there must be a causal link between the considerations and the reading the court adopts. The four classical interpretive canons operate at the level of these considerations: grammatical, systematic, historical, and objective-teleological interpretation. There has long been consensus that there is no consensus on the precise definitions of the interpretive canons \citep{Kriele1967}. The following definitions are based on the account given by Larenz:

Grammatical interpretation is interpretation according to the meaning of an expression or a combination of words in general usage or, if such a usage can be identified, in the special usage of the statutory provision in question \citep[p.~321]{Larenz1991}.

Systematic interpretation requires first and foremost attention to context, as is necessary for understanding any connected speech or writing. Beyond this, it refers to the substantive consistency of provisions within a single body of rules, and further to attention to the outward arrangement of the statute \citep[up until here][p.~328]{Larenz1991} as well as to provisions lying outside the statute that are relevant to its understanding.

Historical interpretation is interpretation according to the legislator's regulatory intent \citep[p.~328]{Larenz1991}.\footnote{On Larenz's account, the regulatory intent consists, in particular, of all those considerations that remained unchallenged in the deliberations. He argues that the view of an individual ministerial official or politician does not permit an inference as to the legislator's regulatory intent. While this is plausible, this part of the definition was nevertheless not adopted into the definition used for this study. Only in the rarest of cases will it be apparent to an LLM from the text of the decision itself whether a consideration remained unchallenged in the deliberations. Furthermore, the definition would then no longer merely capture the occurrence of a particular interpretive canon as a phenomenon, but would also entail a value judgment as to whether the court conducted the interpretation in the individual case in a methodologically sound way. Historical considerations that would commonly be subsumed under the term would thus run the risk of being left out of account.}

Objective-teleological interpretation means the interpretation of a statutory provision according to the structures of the regulated subject area and the underlying legal-ethical principles \citep[up until here][p.~328]{Larenz1991}, without recourse to the legislator's regulatory intent.

\section{From Definition to a Sentence-Level Task}
\label{sec:pipeline}

The criteria of \cref{sec:task} do not by themselves fix how a decision text is to be broken into units for classification. This section motivates the unit of analysis and describes how the subtasks compose into a multi-stage pipeline; the annotation scheme of \cref{sec:dataset} and the structure of the benchmark both follow from it.

\subsection{Unit of Analysis}
\label{sec:unit}

Ideally, one would extract each reading and each supporting consideration at its exact word boundaries, which need not coincide with sentence or paragraph boundaries. The drawbacks of such free-span extraction become apparent in an example:

\begin{example}\label{ex:bremurlg}
``When the Bremen Leave Act was enacted and amended, postal employees were members of the administration of the United Economic Area. That the Bremen Leave Act was, according to the will of the Bremen legislator, meant to extend to them as well was not expressly stated in the deliberations on the statute; it follows, however, beyond doubt from Section~1 BremUrlG. Under that provision, the Act applies `\ldots\ to the administrations and enterprises of the public service that have their seat in the Land of Bremen or operate in the Land of Bremen'. In the deliberations it was expressly emphasized that the Leave Act was to apply to `all employees of the Bremen state territory' [\ldots]. For the delimitation of the class of persons entitled, the place of service was thus to be decisive, and not the identity of the employer.'' (BVerfGE 11, 89 (95))
\end{example}

The reading lies in the second sentence and concerns the personal scope of Section~1 BremUrlG (\emph{Bremisches Urlaubsgesetz}, the Bremen Leave Act). A model instructed to find all considerations supporting this reading, and to name the canons used in each, might point to the penultimate sentence, which draws on the parliamentary deliberations, and correctly identify a historical argument. Yet the third sentence also carries a grammatical argument, since it invokes the wording of the provision. Because the task was to find \emph{all} uses of the canons in the passage, the answer would be incomplete and therefore wrong, and, more importantly, the implicit decision the model took against every other span in the passage cannot be checked. Just as not every use of a canon catches a human reader's eye on a cursory pass, LLMs often overlook arguments when asked to extract an unbounded number of them from a passage. We therefore pose the task so that only a fixed set of answers is possible: every sentence is classified against every criterion it reaches, with missed arguments treated as false negatives.

Choosing the sentence as the unit of classification does not presuppose that an interpretive argument typically fits within a single sentence. The annotated data show the opposite. A single reading is frequently supported by several argument sentences: across all annotated positives, between roughly a third (systematic, historical) and nearly half (grammatical, objective-teleological) of the readings are supported by more than one annotated argument sentence, and the same holds for the argument gate of \cref{sec:dataset}. The sentence is thus a unit of \emph{measurement}, chosen because it captures the components that make up an argument concisely and exhaustively. Reconstructing full arguments would proceed by clustering classified sentences that support the same reading and contain the same canon; this additional step is outside the scope of this paper.

The structure of the dataset does not preclude coarser granularities. Paragraph markers are inherited from the L.L.Con corpus (\cref{sec:dataset}), so the dataset can also be used for paragraph-level classification. The results, however, would not be comparable to the ones reported here, as they would not share the same unit of measurement.

\subsection{Pipeline}
\label{sec:pipeline-stages}

The subtasks compose into a cascading pipeline that mirrors the order of \cref{sec:task}. Every sentence of the reasons is evaluated on the first reading criterion, and each subsequent criterion is asked only where the preceding requirements are satisfied; sentences that pass all three are treated as readings. Relations between sentences can then be viewed as a matrix whose entries evaluate a (reading sentence, candidate argument sentence) pair. Since evaluating all pairs grows quadratically with decision length, candidates are restricted to a window around each reading, with five sentences before and seven after, asymmetric because supporting sentences follow their reading far more often than they precede it (\cref{sec:unit}); the window covers about 90\% of annotated argument pairs, and positives annotated outside it are retained. Each pair is classified on whether the candidate is an argument for the reading at all and, conditional on that, on each of the four canons.

In this paper, the pipeline defines the annotation scheme and the structure of the benchmark: each stage is evaluated in isolation, so that stage-level scores are not confounded by upstream errors. Running the pipeline end-to-end over full decisions, the setting in which upstream errors propagate, is supported by the decision-level release (\cref{sec:dataset}) but not pursued here.

\section{Dataset}
\label{sec:dataset}

\paragraph{Source and scope.}
The decisions are drawn from the L.L.Con corpus of decisions of the German Federal Constitutional Court \citep{Moellers2023}. We draw 28 decisions, all from the official collection (\emph{BVerfGE}). Within each decision, annotation is restricted to the reasons (\emph{Entscheidungsgründe}); the L.L.Con metadata already delimit them at paragraph level and we adopt these boundaries. Reasons are split into sentences with \texttt{distilbert-SBD-de-judgements}, a DistilBERT model fine-tuned for sentence boundary detection in German legal text \citep{brugger2023multilegalsbd}. The detector occasionally emits fragments that are not classifiable sentences, such as section headings and bare enumerators (``II.'', ``B.'', ``B.-I.'') that carry no assertion; we mark these as non-evaluable and exclude them from the dataset before annotation and classification, so they are never emitted as instances on any subtask.\footnote{Operationally, a span is treated as non-evaluable if it contains no alphabetic character or is at most five characters long. This matters most for \texttt{nicht\_abstrakt}, since a bare enumerator trivially satisfies ``not abstract'' and leaving these spans in would inflate that subtask's positive class.} The sentence is the unit of annotation and classification throughout. Of the 28 decisions, 15 are annotated \emph{exhaustively}, with every sentence of the reasons annotated, and the remaining 13 decisions are annotated \emph{selectively}, with only the positive occurrences of the interpretive canons recorded. In all 28 decisions, every annotated sentence is annotated along the cascade of \cref{sec:pipeline-stages}, receiving a label on each subtask it reaches; sentences that fail an earlier gate are not labeled further, since being able to reliably assign a sensible later label is conditional on the preceding requirements in the cascade being satisfied. This means that negatives for any class are only drawn from sentences explicitly annotated as negative for the class in question, and never drawn from those which have not received a label at all. It also means that the unannotated sentences in the selectively annotated decisions are not included in this dataset and are not, and should not be, treated as negatives. For the abstract and self-advanced criteria, positives and negatives come from the exhaustive set alone. The statutory reference subtask also draws positives from selectively annotated decisions, while its negatives come only from the exhaustive set. For the argument subtasks, selectively annotated decisions also contribute negatives. 

\paragraph{Subtasks.}
The annotation scheme follows the steps laid out in \cref{sec:task}.
Throughout, we name each subtask by its English concept and give in
parentheses the identifier under which it appears in the released data and
in the result tables below; the identifiers are German, following the
annotation scheme and the language of the source material.
Each candidate sentence is first scored on the three reading sub-criteria
that remain operationalizable on the basis of the decision text alone:
\textit{not abstract} (\texttt{nicht\_abstrakt}), \textit{not self-advanced}
(\texttt{nicht\_selbst\_aufgestellt}), and the \textit{statutory reference}
(\texttt{konkretes\_gesetz}), which asks for the provision whose content the
sentence determines and so decides both halves of \cref{sec:reference-point}
in a single pass.\footnote{The first two criteria \emph{abstract} and \emph{self-advanced} are framed as negations of how they were presented in \cref{sec:reading}; for the rationale, see the section on class sizes below.} Sentences that pass all three are
treated as readings; the sentences in the window described in \cref{sec:pipeline-stages} are then evaluated as
\emph{candidate arguments}, first on whether the candidate supports the
reading at all, i.e., the \textit{argument gate} (\texttt{argument}), and then,
conditional on that, on each of the
four canons of \textit{grammatical} (\texttt{wortlaut}), \textit{systematic} (\texttt{systematik}),
\textit{historical} (\texttt{geschichte}) and \textit{objective-teleological} (\texttt{zweck}) interpretation.
The gate is not a fifth canon: it asks \emph{whether} a sentence supports the
reading, where the canons ask \emph{which kind} of support it offers, and the
four canons do not exhaust the ways a court can argue. A sentence can
therefore pass the gate and still be negative on all four canons.
This gives eight subtasks in total: seven binary
classification subtasks, plus the statutory reference task, whose label is
a list of all the statutory provisions that the claim of the sentence refers to; usually, this will either be a single provision or no provision at all. Argument subtasks are evaluated on a \emph{sentence pair}, i.e. a
candidate argument sentence together with the reading it is meant to
support, since canon labels depend on
what is being argued for (cf.\ \cref{ex:redetermination,ex:electoral-act} in \cref{sec:reading}).
\paragraph{Class sizes.}
\label{sec:class-sizes}
Each of the three reading sub-criteria is represented by 400 sentences
(100 positive / 300 negative). Each of the five argument subtasks is
represented by 200 sentence pairs (50 positive / 150 negative). The
argument subtasks are smaller because two filters compound: the pool of
valid readings is itself a bottleneck, and many of the readings that
do qualify are not supported by any explicit argument in the surrounding
text. For the rarer argument classes, the 1:3 positive--negative ratio is intentionally more favorable than the much lower true base rate to prevent class imbalances. For the abstract and self-advanced criteria, the natural ratio is closer to 3:1, which is why the task is framed inversely here, i.e.\ whether an assertion is \emph{not} abstract or \emph{not} self-advanced. The negatives of the \textit{statutory reference} subtask come in two kinds, corresponding to the two halves of \cref{sec:reference-point}: a sentence may name no specific provision at all, or name one without determining its content. Because the first kind (\texttt{keine\_konkrete\_gesetzesbestimmung}) is concentrated in a few decisions, a decision-disjoint draw clusters it, so each split preserves the pool's proportions of the two kinds up to integer rounding.\footnote{The exact split used is 78:22. Of the 300 negatives, 234 are of the second kind (\texttt{kein\_bestimmter\_inhalt}) and 66 of the first, allocated per split as 47/13, 47/13, and 140/40 across train, validation, and test.}

All splits are stratified to the class ratio of 1:3 described above. Furthermore, they are decision-disjoint throughout, and a decision quoted as a worked example in a subtask's prompt is additionally held out of that subtask's test split, so that no test sentence is one the prompt has already shown with its label.\footnote{The decisions held out of each subtask's test split are: \texttt{BVerfGE61,126} and \texttt{cs20090421\_2bvc000206} for \texttt{konkretes\_gesetz}; \texttt{BVerfGE57,170} for \texttt{systematik}, for \texttt{geschichte}, and for \texttt{nicht\_selbst\_aufgestellt}; and \texttt{BVerfGE57,170}, \texttt{BVerfGE61,126}, and \texttt{BVerfGE62,338} for \texttt{nicht\_abstrakt}.}

All annotations were produced by a single annotator who pursues a PhD in law. Each instance carries a free-text \texttt{reasoning} field that allows for more detailed explanations of the annotation in edge cases. One subtask we deliberately do not provide as a stand-alone classification target yet is whether a candidate sentence is an \emph{assertion}, as opposed to a verbatim
restatement of the statute: deciding this reliably requires comparing the
sentence against the statutory text in force at the time of the decision,
and the historical statutory text is not yet available at scale for
Germany. The related challenge of distinguishing a court's own assertion
from a reference to a statutory provision has recently been taken up for
French law by \citet{holzenberger2026}. We instead flag sentences that
look like potential restatements in the instance-level data so they can
be re-checked once historical statutory data for Germany becomes
available.

\paragraph{Release.}
The dataset is released at publication in two complementary forms.
\textit{Decision-level}: one record per fully annotated decision, pairing
the plain text with JSON metadata that encodes every annotation as
character offsets into that text along with paragraph-level structural
metadata inherited from L.L.Con. This format is intended for end-to-end
evaluation that runs the full pipeline over a whole decision, which is not pursued in this paper. 
\textit{Instance-level}: one record per (subtask, candidate) pair,
grouped by subtask and split. Each record carries the candidate
sentence, the surrounding context window with the candidate marked by
inline XML tags (\texttt{<deutung>}/\texttt{<potential\_argument>};
\emph{Deutung} is the German term for a reading),
and the gold label with its reasoning. For the statutory reference subtask,
a wider context window is added because the relevant statute citation
often lies further upstream than for the other criteria.

\section{Baseline Evaluations}
\label{sec:experiments}

We evaluate four LLMs from three model families under different prompts to establish a first baseline performance for the introduced task: DeepSeek-V4 in its Pro and Flash variants, Gemini~3.5 Flash Lite, and MiniMax~M3. DeepSeek-V4 and MiniMax~M3 are open-weight models; Gemini~3.5 Flash Lite is a proprietary model. All four are reasoning models and were run at temperature 1.0.

The models are compared across two sets of prompts: one that is not model-specific, crafted by an expert, and one that is optimized using Genetic-Pareto (GEPA) \citep{agrawal2025gepa}, starting from the expert prompt.\footnote{The expert prompts illustrate each criterion with worked examples. Most are excerpts from the corpus, but some are constructed to isolate a single distinction as cleanly as a real passage rarely does; these carry placeholder citations such as ``BVerfGE 123, 456'' and are not references to decisions. All prompts are released with the datasets (\cref{fn:wandb}).} The purpose of the GEPA condition is to validate the expert baseline: at a time when automatic prompt optimization frequently outperforms hand-written prompts, it is useful to assess whether optimization substantially improves on a hand-crafted baseline. GEPA thus serves as an orientation point for what current models can achieve on the task, against which the expert prompt is anchored. In the GEPA condition, the models are accordingly not evaluated across one static prompt, but across a model-specific prompt produced by an optimizer with static settings: DeepSeek-V4-Pro was used as the reflection language model for all four evaluated models, at temperature 1.0 with a maximum of 8,000 tokens, and the maximum metric calls budget was set to auto=light. The built-in instruction proposer was modified so that the resulting prompt would be German.

GEPA \citep{agrawal2025gepa} evolves prompts by reflecting in natural
language on execution traces and selecting along a Pareto frontier of
candidates; because it learns from a handful of rollouts rather than
from gradient updates over many labeled examples, it is markedly more
sample-efficient than tuning-based adaptation and the optimization
stage consumes correspondingly little data. We optimize the binary subtasks for accuracy (though we report F1, \cref{app:more-models}), and the statutory reference task for sample-averaged F1. Because GEPA scores each rollout individually, its target must be a per-instance metric: a single binary decision admits only correct/incorrect (accuracy in aggregate), whereas the statutory reference output is a set of provisions with a per-document F1 to average. We exploit GEPA's efficiency with an atypical 20/20/60 split: for each subtask, 20\% of instances drive GEPA's reflective search, a further 20\% provide the validation signal it optimizes against, and the remaining 60\% are held out as a test set touched only once, for the numbers reported here. Allocating the majority of the labeled data to the test split yields the most reliable evaluation our annotation budget allows. 

At classification time the model never sees a unit in isolation. Each instance embeds the unit under classification in its surrounding sentence context, with the target span(s) marked by inline XML tags (\texttt{<deutung>} for a reading, \texttt{<potential\_argument>} for a candidate argument); the unit is a single sentence for the reading criteria and a (reading, candidate) pair for the argument subtasks. The context spans two sentences on either side, widened to the ten preceding sentences for the statutory reference subtask, whose governing citation often lies further upstream (\cref{sec:dataset}).

Every subtask is binary except for the statutory reference subtask, which is a citation-extraction target. For each instance the model first emits a free-text \texttt{reasoning} field and then a structured decision: a binary value for the seven binary subtasks, or a list of statutory provisions for the statutory reference subtask. The output is produced by constrained decoding wherever the backbone supports it, and falls back to JSON mode otherwise.

For the seven binary subtasks we report $F_1$ on the positive class; per-subtask precision and recall are given in \cref{app:more-models}. The \textit{statutory reference} subtask is reported separately because it is scored by set-overlap $F_1$ between predicted and gold statutory citations rather than as a binary decision, so a single binary $F_1$ would not be comparable.

Deciding whether a predicted citation matches a gold is challenging by string comparison, because the same provision admits many surface forms (\textit{\S~2 Rechtshilfegesetz} / \textit{\S~2 des Rechtshilfegesetzes} / \textit{\S~2 RhG}). We therefore match with an LLM judge (DeepSeek-V4-Pro), which is shown the gold list for one sentence together with a single predicted citation and maps that prediction to at most one entry of the list, or to none. The judge is instructed to tolerate formatting variation, for example roman numerals for \textit{Abs.}, a trailing number for \textit{Satz}, alternative statutory abbreviations, and differences in whitespace, punctuation and case, but to insist on the exact section, article, paragraph, number and lettered subdivision. Predicted citations are deduplicated before judging; a prediction that the judge matches to no gold entry, or to one already claimed by an earlier prediction, counts as a false positive, and every gold entry left unclaimed counts as a false negative. Per-document set-overlap $F_1$ averages the resulting per-sentence $F_1$ over the test sentences, crediting $1.0$ where gold and prediction are both empty; span-level precision, recall and $F_1$ pool the counts over all sentences instead. Because the judge is itself a language model, its decisions are released alongside the predictions, so that the reported scores can be recomputed exactly without re-running it. The judge prompt is released with the other prompts.

All intervals are 95\% bootstrap confidence intervals from 1{,}000 instance-level resamples of the test set, reported as the 2.5th and 97.5th percentiles. These intervals do not account for dependence between instances from the same decision.

\begin{table*}[t]
\centering
\small
\caption{Per-subtask $F_1$ on the positive class for all four models under the expert and GEPA prompts. Point estimates only; 95\% bootstrap CIs and per-cell precision and recall are in \cref{app:more-models}. ``Mean'' averages the seven binary subtasks.}
\label{tab:main-results}
\begin{tabular}{lcccccccc}
\toprule
 & \multicolumn{2}{c}{DS-V4-Pro} & \multicolumn{2}{c}{DS-V4-Flash} & \multicolumn{2}{c}{Gemini-3.5-FL} & \multicolumn{2}{c}{MiniMax-M3} \\
\cmidrule(lr){2-3}\cmidrule(lr){4-5}\cmidrule(lr){6-7}\cmidrule(lr){8-9}
Subtask & Exp & GEPA & Exp & GEPA & Exp & GEPA & Exp & GEPA \\
\midrule
\multicolumn{9}{l}{\emph{Reading criteria}} \\
\quad \texttt{nicht\_abstrakt} & 73.1 & 71.3 & 71.1 & 75.4 & 73.4 & 71.4 & 80.6 & 74.0 \\
\quad \texttt{nicht\_selbst\_aufgestellt} & 94.3 & 94.3 & 94.3 & 93.4 & 95.7 & 94.9 & 94.2 & 93.4 \\
\addlinespace
\multicolumn{9}{l}{\emph{Argument identification}} \\
\quad \texttt{argument} (gate) & 65.7 & 68.7 & 74.0 & 64.4 & 73.2 & 73.0 & 74.7 & 79.4 \\
\multicolumn{9}{l}{\quad\emph{Canons:}} \\
\quad\quad \texttt{wortlaut} & 86.7 & 89.2 & 78.1 & 82.0 & 79.4 & 78.7 & 83.9 & 85.2 \\
\quad\quad \texttt{systematik} & 70.2 & 64.4 & 63.5 & 47.9 & 47.8 & 47.1 & 55.6 & 70.8 \\
\quad\quad \texttt{geschichte} & 62.1 & 71.2 & 62.0 & 66.7 & 72.7 & 84.7 & 64.6 & 77.4 \\
\quad\quad \texttt{zweck} & 74.1 & 66.7 & 71.7 & 62.7 & 66.7 & 77.8 & 73.1 & 74.1 \\
\midrule
Mean (7 binary) & 75.2 & 75.1 & 73.5 & 70.4 & 72.7 & 75.4 & 75.2 & 79.2 \\
\midrule
\texttt{konkretes\_gesetz}$^{\dagger}$ & 74.7 & 78.6 & 77.2 & 76.7 & 64.1 & 68.6 & 74.2 & 74.7 \\
\bottomrule
\end{tabular}
\vskip 2pt
{\footnotesize\raggedright $^{\dagger}$\textit{Statutory reference} is a citation-extraction target scored by per-document set-overlap $F_1$, not a binary $F_1$; its much lower span-level $F_1$ is discussed in \cref{sec:results}.\par}
\end{table*}

\section{Results}
\label{sec:results}

\cref{tab:main-results} reports $F_1$ scores for every run at the subtask level; more detailed scores can be found in the Appendix in \crefrange{tab:pro-results-app}{tab:statref-app}.

\paragraph{No model family dominates.}
Averaged over the seven binary subtasks, the four models cluster between
70.4 and 79.2 $F_1$, a spread of under nine points. MiniMax M3
under GEPA is nominally highest (79.2) and DeepSeek-V4-Flash under
GEPA nominally lowest (70.4), but no model consistently leads across subtasks. We therefore read the table as a set of baselines
rather than a ranking.

\paragraph{The expert baseline holds up against optimized prompts.}
Under the tested configuration, the GEPA prompts do not systematically beat the expert prompt: on
the binary mean, GEPA is ahead for Gemini 3.5 Flash Lite (75.4 vs.\ 72.7)
and MiniMax M3 (79.2 vs.\ 75.2), behind for DeepSeek-V4-Flash (70.4 vs.\
73.5), and level for DeepSeek-V4-Pro (75.1 vs.\ 75.2). The expert prompt remains a meaningful baseline: its mean binary $F_1$ does not sit far below what the tested GEPA configuration attains.

\paragraph{The statutory-reference scores need separate reading.}
The headline set-overlap $F_1$ on this subtask (64.1--78.6) is a
per-document metric dominated by the 180 of 240 test sentences whose gold citation set is empty: a model that correctly abstains on those scores $1.0$. At the level of individual citations the extraction is far weaker, with span-level $F_1$ only 28.4--42.4 and precision between 21 and 47 (\cref{tab:statref-app}). Both prompt conditions over-extract, emitting a citation for sentences that merely mention or evaluate a provision without asserting its content, so the low precision is a property of the task rather than of one prompt.

\paragraph{Canon difficulty is stable at its extremes and plausibly holds up against legal intuition.}
The classifiers achieve the best scores on grammatical interpretation in seven of the eight model--prompt conditions (mean 82.9), and the lowest ones on systematic interpretation in six of eight (mean 58.4); historical and objective-teleological interpretation sit between and are effectively tied (means 70.2 and 70.9). This is at least a plausible result as the grammatical canon has a clear anchor in the boundary of the word, whereas the reach of the systematic canon is contested and the boundaries of purpose are blurry. Systematic interpretation's low score comes from failing in both directions at once, over- and under-attributing the canon (\cref{sec:error-analysis}).

\section{Error Analysis}
\label{sec:error-analysis}

\paragraph{Multi-sentence arguments drive the false negatives.}
We annotate every sentence of an argument that runs across
several sentences; models miss continuing or concluding sentences of such arguments that lack canon-specific cues. This is why objective-teleological interpretation, whose arguments are often built up gradually, loses ground on recall.

\paragraph{Shallow cues drive the false positives.}
On the argument gate the models accept restatements and mere consequences of the reading as though they were grounds; on the remaining subtasks they latch onto a surface proxy, for example the token ``Wortlaut'' (grammatical), the citation of any distinct provision (systematic), or any section number (statutory reference, hence its over-extraction), none of which is sufficient for the criterion it stands in for. Systematic interpretation is the canon with the lowest scores likely because its proxy misfires both ways, firing on any parallel citation and missing systematic arguments that name no second provision.

\section{Discussion and Limitations}
\label{sec:discussion}

Several limitations bound our results. All gold labels come from a single annotator, which yields consistent application of the criteria in \cref{sec:task} but does not allow for a measure of inter-annotator agreement. The corpus is drawn entirely from one court in one language, so we make no claim of transfer to other jurisdictions or legal traditions. Furthermore, while the baseline covers four models from three families, it remains unclear how well the most capable closed-source models would close the remaining gap. The confidence intervals in \cref{sec:results} are wide, a direct effect of the small test sets the current dataset affords, especially for the rarer argument subtasks; differences between prompt conditions should not be over-interpreted. Extending the corpus using multiple annotators, which we intend to do, is the most direct remedy.

\section{Conclusion}
\label{sec:conclusion}

This paper set out to make the classification of interpretive canons tractable at the sentence level, rather than the paragraph that prior work has relied on. We made three contributions: we operationalized the four classical canons as articulated by Larenz in the tradition of Savigny; we released an expert-annotated dataset of German Federal Constitutional Court decisions, exhaustively labeled for fifteen decisions and selectively extended for rarer classes; and we reported baseline evaluations of four LLMs from three model families under expert-written and GEPA-optimized prompts. Two findings stand out. First, the relative difficulty of the canons is at least plausible in light of legal methodology, with grammatical interpretation usually the easiest canon to identify and systematic interpretation usually the hardest. Second, prompts optimized with the tested GEPA configuration do not systematically outperform the expert hand-written prompts in mean binary $F_1$, supporting the reported baseline as a meaningful reference point. Given the wide confidence intervals and the modest size of the current dataset, we read these results as a starting point that invites extending the corpus and building toward better evaluations of interpretive canons.

\section*{Impact Statement}
This paper presents work whose goal is to advance the field of Machine Learning. There are many potential societal consequences of our work, none of which we feel must be specifically highlighted here.

\section*{Acknowledgments}
I thank Professor Andreas Engert for his support throughout this work, and the anonymous reviewers of the ICML 2026 Workshop on AI for Law for their constructive feedback. Any remaining errors are my own.

\bibliography{references}
\bibliographystyle{icml2026}           

\newpage
\appendix
\onecolumn
\raggedbottom
\makeatletter
\setlength{\@fptop}{0pt}            
\setlength{\@fpsep}{8pt}            
\setlength{\@fpbot}{0pt plus 1fil}  
\makeatother

\section{Complete Model Scores}
\label{app:more-models}

Tables~\ref{tab:pro-results-app}--\ref{tab:minimax-results} report
per-subtask precision, recall and $F_1$ for all four models on the seven
binary subtasks, and \cref{tab:statref-app} reports the \textit{statutory
reference} subtask. Subtasks are named by the identifiers introduced in
\cref{sec:dataset}. Precision and recall are point estimates; $F_1$ carries
a 95\% bootstrap CI (1{,}000 resamples). ``Mean'' averages the seven binary
subtasks.

\begin{table}[!ht]
\centering
\small
\caption{Precision, recall and $F_1$ (positive class) for DeepSeek-V4-Pro across the seven binary subtasks. P and R are point estimates; $F_1$ carries the 95\% bootstrap CI (1{,}000 resamples). \textit{Statutory reference} is reported separately in \cref{tab:statref-app}.}
\label{tab:pro-results-app}
\begin{tabular}{l ccc ccc}
\toprule
 & \multicolumn{3}{c}{Expert} & \multicolumn{3}{c}{GEPA} \\
\cmidrule(lr){2-4}\cmidrule(lr){5-7}
Subtask & P & R & $F_1$ (95\% CI) & P & R & $F_1$ (95\% CI) \\
\midrule
\multicolumn{7}{l}{\emph{Reading criteria}} \\
\quad \texttt{nicht\_abstrakt} & 62.4 & 88.3 & 73.1~(64.4--80.3) & 66.7 & 76.7 & 71.3~(61.8--79.5) \\
\quad \texttt{nicht\_selbst\_aufgestellt} & 92.1 & 96.7 & 94.3~(89.4--97.9) & 92.1 & 96.7 & 94.3~(89.4--97.9) \\
\multicolumn{7}{l}{\emph{Argument gate}} \\
\quad \texttt{argument} & 59.5 & 73.3 & 65.7~(51.9--76.9) & 62.2 & 76.7 & 68.7~(55.6--80.0) \\
\multicolumn{7}{l}{\emph{Canons}} \\
\quad \texttt{wortlaut} & 86.7 & 86.7 & 86.7~(76.7--94.3) & 82.9 & 96.7 & 89.2~(78.8--96.4) \\
\quad \texttt{systematik} & 74.1 & 66.7 & 70.2~(55.0--82.1) & 65.5 & 63.3 & 64.4~(47.8--77.3) \\
\quad \texttt{geschichte} & 64.3 & 60.0 & 62.1~(45.5--75.4) & 72.4 & 70.0 & 71.2~(56.0--82.8) \\
\quad \texttt{zweck} & 83.3 & 66.7 & 74.1~(58.3--85.7) & 75.0 & 60.0 & 66.7~(50.0--80.0) \\
\midrule
Mean (7 binary) & 74.6 & 76.9 & 75.2 & 73.8 & 77.2 & 75.1 \\
\bottomrule
\end{tabular}
\end{table}

\begin{table}[!ht]
\centering
\small
\caption{Precision, recall and $F_1$ (positive class) for DeepSeek-V4-Flash across the seven binary subtasks. P and R are point estimates; $F_1$ carries the 95\% bootstrap CI (1{,}000 resamples). \textit{Statutory reference} is reported separately in \cref{tab:statref-app}.}
\label{tab:flash-results}
\begin{tabular}{l ccc ccc}
\toprule
 & \multicolumn{3}{c}{Expert} & \multicolumn{3}{c}{GEPA} \\
\cmidrule(lr){2-4}\cmidrule(lr){5-7}
Subtask & P & R & $F_1$ (95\% CI) & P & R & $F_1$ (95\% CI) \\
\midrule
\multicolumn{7}{l}{\emph{Reading criteria}} \\
\quad \texttt{nicht\_abstrakt} & 59.6 & 88.3 & 71.1~(62.1--78.8) & 66.7 & 86.7 & 75.4~(66.2--82.9) \\
\quad \texttt{nicht\_selbst\_aufgestellt} & 92.1 & 96.7 & 94.3~(89.4--97.9) & 91.9 & 95.0 & 93.4~(88.4--97.6) \\
\multicolumn{7}{l}{\emph{Argument gate}} \\
\quad \texttt{argument} & 62.8 & 90.0 & 74.0~(61.5--84.1) & 65.5 & 63.3 & 64.4~(48.0--77.3) \\
\multicolumn{7}{l}{\emph{Canons}} \\
\quad \texttt{wortlaut} & 73.5 & 83.3 & 78.1~(65.1--88.0) & 80.6 & 83.3 & 82.0~(69.1--91.3) \\
\quad \texttt{systematik} & 60.6 & 66.7 & 63.5~(47.8--76.5) & 41.5 & 56.7 & 47.9~(31.6--61.4) \\
\quad \texttt{geschichte} & 53.7 & 73.3 & 62.0~(47.9--74.3) & 59.0 & 76.7 & 66.7~(52.8--77.8) \\
\quad \texttt{zweck} & 82.6 & 63.3 & 71.7~(56.4--84.4) & 76.2 & 53.3 & 62.7~(43.9--76.6) \\
\midrule
Mean (7 binary) & 69.3 & 80.2 & 73.5 & 68.8 & 73.6 & 70.4 \\
\bottomrule
\end{tabular}
\end{table}

\begin{table}[!ht]
\centering
\small
\caption{Precision, recall and $F_1$ (positive class) for Gemini 3.5 Flash Lite across the seven binary subtasks. P and R are point estimates; $F_1$ carries the 95\% bootstrap CI (1{,}000 resamples). \textit{Statutory reference} is reported separately in \cref{tab:statref-app}.}
\label{tab:gemini-results}
\begin{tabular}{l ccc ccc}
\toprule
 & \multicolumn{3}{c}{Expert} & \multicolumn{3}{c}{GEPA} \\
\cmidrule(lr){2-4}\cmidrule(lr){5-7}
Subtask & P & R & $F_1$ (95\% CI) & P & R & $F_1$ (95\% CI) \\
\midrule
\multicolumn{7}{l}{\emph{Reading criteria}} \\
\quad \texttt{nicht\_abstrakt} & 64.6 & 85.0 & 73.4~(64.4--80.6) & 76.9 & 66.7 & 71.4~(61.3--80.3) \\
\quad \texttt{nicht\_selbst\_aufgestellt} & 98.2 & 93.3 & 95.7~(91.4--99.1) & 96.6 & 93.3 & 94.9~(90.6--98.6) \\
\multicolumn{7}{l}{\emph{Argument gate}} \\
\quad \texttt{argument} & 63.4 & 86.7 & 73.2~(60.3--83.6) & 69.7 & 76.7 & 73.0~(59.3--84.2) \\
\multicolumn{7}{l}{\emph{Canons}} \\
\quad \texttt{wortlaut} & 75.8 & 83.3 & 79.4~(66.7--88.6) & 77.4 & 80.0 & 78.7~(66.7--88.6) \\
\quad \texttt{systematik} & 43.2 & 53.3 & 47.8~(32.0--62.5) & 57.1 & 40.0 & 47.1~(27.3--64.0) \\
\quad \texttt{geschichte} & 80.0 & 66.7 & 72.7~(58.2--84.4) & 86.2 & 83.3 & 84.7~(74.3--93.5) \\
\quad \texttt{zweck} & 81.0 & 56.7 & 66.7~(50.0--80.6) & 87.5 & 70.0 & 77.8~(63.6--88.9) \\
\midrule
Mean (7 binary) & 72.3 & 75.0 & 72.7 & 78.8 & 72.9 & 75.4 \\
\bottomrule
\end{tabular}
\end{table}

\begin{table}[!ht]
\centering
\small
\caption{Precision, recall and $F_1$ (positive class) for MiniMax M3 across the seven binary subtasks. P and R are point estimates; $F_1$ carries the 95\% bootstrap CI (1{,}000 resamples). \textit{Statutory reference} is reported separately in \cref{tab:statref-app}.}
\label{tab:minimax-results}
\begin{tabular}{l ccc ccc}
\toprule
 & \multicolumn{3}{c}{Expert} & \multicolumn{3}{c}{GEPA} \\
\cmidrule(lr){2-4}\cmidrule(lr){5-7}
Subtask & P & R & $F_1$ (95\% CI) & P & R & $F_1$ (95\% CI) \\
\midrule
\multicolumn{7}{l}{\emph{Reading criteria}} \\
\quad \texttt{nicht\_abstrakt} & 75.4 & 86.7 & 80.6~(72.2--87.1) & 70.1 & 78.3 & 74.0~(64.6--81.7) \\
\quad \texttt{nicht\_selbst\_aufgestellt} & 93.4 & 95.0 & 94.2~(89.7--98.1) & 91.9 & 95.0 & 93.4~(88.4--97.4) \\
\multicolumn{7}{l}{\emph{Argument gate}} \\
\quad \texttt{argument} & 62.2 & 93.3 & 74.7~(62.5--84.1) & 71.1 & 90.0 & 79.4~(67.6--88.6) \\
\multicolumn{7}{l}{\emph{Canons}} \\
\quad \texttt{wortlaut} & 81.2 & 86.7 & 83.9~(72.0--92.5) & 83.9 & 86.7 & 85.2~(74.3--93.8) \\
\quad \texttt{systematik} & 62.5 & 50.0 & 55.6~(37.7--70.4) & 65.7 & 76.7 & 70.8~(55.3--81.6) \\
\quad \texttt{geschichte} & 60.0 & 70.0 & 64.6~(49.2--76.7) & 75.0 & 80.0 & 77.4~(64.6--87.0) \\
\quad \texttt{zweck} & 86.4 & 63.3 & 73.1~(57.1--84.6) & 83.3 & 66.7 & 74.1~(58.3--86.2) \\
\midrule
Mean (7 binary) & 74.4 & 77.9 & 75.2 & 77.3 & 81.9 & 79.2 \\
\bottomrule
\end{tabular}
\end{table}

\begin{table}[!ht]
\centering
\small
\caption{\textit{Statutory reference} (\texttt{konkretes\_gesetz}), the citation-extraction subtask. \emph{Set-overlap} $F_1$ is the per-document metric reported in the main text and averages over the 180/240 empty-gold rows (a model that correctly returns no citation scores 1.0 on such a row). \emph{Span-level} P/R/$F_1$ count individual predicted vs.\ gold citations and expose the over-extraction directly. $F_1$ columns carry 95\% bootstrap CIs.}
\label{tab:statref-app}
\begin{tabular}{ll c ccc}
\toprule
 & & Set-overlap & \multicolumn{3}{c}{Span-level} \\
\cmidrule(lr){4-6}
Model & Prompt & $F_1$ (95\% CI) & P & R & $F_1$ (95\% CI) \\
\midrule
DeepSeek-V4-Pro & Expert & 74.7~(69.2--80.3) & 36.8 & 41.7 & 39.1~(28.3--49.6) \\
 & GEPA & 78.6~(73.2--83.8) & 46.9 & 25.0 & 32.6~(20.9--44.7) \\
\addlinespace
DeepSeek-V4-Flash & Expert & 77.2~(71.9--82.4) & 39.7 & 45.0 & 42.2~(30.1--52.9) \\
 & GEPA & 76.7~(71.0--81.7) & 38.9 & 46.7 & 42.4~(31.1--52.9) \\
\addlinespace
Gemini 3.5 Flash Lite & Expert & 64.1~(58.1--70.0) & 20.8 & 45.0 & 28.4~(20.2--36.9) \\
 & GEPA & 68.6~(62.6--74.6) & 27.7 & 55.0 & 36.9~(28.3--46.0) \\
\addlinespace
MiniMax M3 & Expert & 74.2~(68.1--79.8) & 35.2 & 51.7 & 41.9~(31.5--52.2) \\
 & GEPA & 74.7~(69.2--79.7) & 40.0 & 43.3 & 41.6~(30.8--51.3) \\
\bottomrule
\end{tabular}
\end{table}

\clearpage
\section{Expert System Prompts}
\label{app:prompts}

The eight hand-written expert system prompts are reproduced verbatim below. They
are model-agnostic, identical across the four models. Each operationalizes
one subtask: it states the criterion and illustrates it with worked positive and
negative examples, some constructed with placeholder citations such as
``BVerfGE~123, 456'' rather than real decisions. The paired user prompts, which
wrap the instance's input fields (the context field \texttt{text\_with\_context},
replaced by its wider variant \texttt{text\_with\_context\_konkretes\_gesetz} for
the statutory reference subtask, plus, for the argument subtasks, the reading
sentence, the candidate sentence and the reading's statutory reference), are
omitted, as
are the model-specific GEPA prompts; all prompts are released with the datasets
(\cref{fn:wandb}). The prompts are in German, matching the corpus.

\subsection{Not abstract (\texttt{nicht\_abstrakt})}
\begin{lstlisting}[style=promptstyle]
Das übergreifende Projekt, in dem du genutzt wirst, ist eine Studie, in der untersucht wird, wie sich die Verwendung von Auslegungsmethoden in den Entscheidungen des Bundesverfassungsgerichts im Laufe der Geschichte entwickelt hat. Deine Aufgabe ist Teil eines Workflows, der es zum Ziel hat, einen Satz Metadaten zu erstellen, der für jede Entscheidung genau angibt, an welcher Stelle das Gericht Auslegung mittels der vier Auslegungskriterien betrieben hat.

Auslegung bedeutet, sich für eine unter mehreren möglichen Deutungen einer Gesetzesbestimmung aufgrund von Überlegungen zu entscheiden.

Aus dieser Definition lassen sich die verschiedenen Schritte ableiten, aus denen sich der Workflow zusammensetzt:

1. Deutungen identifizieren
2. Identifizieren, ob es sich bei bestimmten Sätzen überhaupt um solche handelt, die eine Begründung für die Deutung darstellen.
3. Identifizieren, welche Auslegungskriterien in dem Argument verwendet werden.

Du wirst dich ausschließlich mit dem ersten Schritt dieses Workflows befassen.

Eine Deutung ist eine Behauptung, die das Gericht selbst aufstellt und die zum Gegenstand hat, dass eine bestimmte Gesetzesbestimmung abstrakt einen bestimmten Inhalt habe.

Eine Behauptung über den Inhalt einer Gesetzesbestimmung oder eines allgemeinen Rechtsprinzips ist abstrakt, wenn sie Geltung beansprucht nicht nur in Bezug auf den einzigartigen Lebenssachverhalt, der der Entscheidung zugrunde liegt, sondern für unendlich viele Situationen, in denen die Gesetzesbestimmung oder das Rechtsprinzip Anwendung findet.

Deine Aufgabe ist es zu entscheiden, ob eine Behauptung **nicht** abstrakt ist. Der inhaltliche Maßstab ist unverändert der soeben definierte; nur die Richtung der Antwort ist umgekehrt. Das Ausgabefeld `nicht_abstrakt` ist also genau dann `true`, wenn die Behauptung nach diesem Maßstab **nicht** abstrakt ist, und `false`, wenn sie abstrakt ist.

Positive Beispiele (`nicht_abstrakt: true`, die Behauptung ist also nicht abstrakt):

<example>Im vorliegenden Fall ist nicht auszuschließen, daß das Oberlandesgericht bei der Beurteilung der Frage, ob eine konkrete Gefährdung der Anstaltsordnung durch den Brief zu besorgen war, bei Anwendung des dargelegten verfassungsrechtlichen Maßstabs zu einer anderen rechtlichen Würdigung gelangt wäre.</example>
Hier wird auf den vorliegenden Fall und die konkrete Entscheidung des Oberlandesgerichts über einen bestimmten Brief Bezug genommen. Es wird keine allgemeingültige Aussage getroffen, daher liegt keine abstrakte Behauptung vor.

<example>"Die Entscheidung des Oberlandesgerichts ist gemäß § 304 Abs. 4 StPO endgültig."</example>
Hier wird keine abstrakte Aussage über § 304 Abs. 4 StPO getroffen, sondern nur konkret *die* Entscheidung des Oberlandesgerichts anhand der Bestimmung bewertet. Daher liegt keine abstrakte Behauptung vor.

<example>Die Voraussetzung des § 147 StPO lagen zweifelsfrei vor; es gab keinen Grund, den Zentralregisterauszug hiervon auszunehmen.</example>
Hier wird § 147 StPO nicht abstrakt ausgelegt, sondern nur auf den konkreten Sachverhalt angewandt: Es wird festgestellt, dass seine Voraussetzungen im entschiedenen Fall vorlagen. Es wird keine allgemeingültige Aussage getroffen, daher liegt keine abstrakte Behauptung vor.

Negative Beispiele (`nicht_abstrakt: false`, die Behauptung ist also abstrakt):

<negative_example>"Es dient verfassungsrechtlich legitimen Zwecken, die externe Teilung der in § 17 VersAusglG genannten Anrechte (Betriebsrenten aus einer Direktzusage oder Unterstützungskasse) auch über die Wertgrenze des § 14 Abs. 2 Nr. 2 VersAusglG hinaus zu erlauben."</negative_example>
Hier wird eine Aussage über die in § 17 der Vorschrift genannten Anrechte getroffen. Diese Aussage wird hier nicht nur für einen konkreten Lebenssachverhalt getroffen, sondern allgemein für alle Fälle, in denen die Vorschrift zur Anwendung kommen könnte. Somit ist die Behauptung abstrakt im Sinne der Definition.

<negative_example>Wird der Brief eines Untersuchungsgefangenen, wie hier, in sinngemäßer Anwendung des § 108 StPO sichergestellt und der Staatsanwaltschaft zur weiteren Veranlassung zugeleitet, so kann sich der Betroffene mit dem Ziel der Herausgabe des Briefes an die Staatsanwaltschaft wenden, jederzeit auf Entscheidung des zuständigen Richters antragen (§ 98 Abs. 2 StPO) und gegen eine etwaige Beschlagnahme im Vorverfahren Beschwerde einlegen (§ 304 Abs. 1 StPO).</negative_example>
Hier wird allgemeingültig ausgeführt, welche Rechtsbehelfe einem Untersuchungsgefangenen bei der Sicherstellung seines Briefes in sinngemäßer Anwendung des § 108 StPO offenstehen. Die Aussage gilt nicht nur für den konkreten Fall ("wie hier"), sondern für alle derartigen Fälle. Es handelt sich daher um eine abstrakte Behauptung.

<negative_example>Doch muß ein solcher Eingriff dem Grundsatz der Verhältnismäßigkeit entsprechen, der sich bereits aus dem Wesen der Grundrechte selbst ergibt und dem als Element des Rechtsstaatsprinzips Verfassungsrang zukommt (vgl. BVerfGE 19, 342 [347 ff.]; 29, 312 [316]).</negative_example>
Dieser Satz legt den Grundsatz der Verhältnismäßigkeit aus und beansprucht Geltung für alle Grundrechtseingriffe, nicht nur für den des konkreten Falls. Daher ist die Behauptung abstrakt.

Du erhältst eine Eingabe:

- `text_with_context`: ein Auszug aus einer Entscheidung des Bundesverfassungsgerichts. Der konkret zu prüfende Satz ist innerhalb des Auszugs mit den Markierungen `<deutung>...</deutung>` umschlossen; die umgebenden Sätze dienen ausschließlich als Kontext.

Begründe immer deine Antwort, bevor du dich entscheidest. Prüfe, ob der Inhalt des mit `<deutung>` markierten Satzes abstrakt im Sinne des Maßstabs aus Definition und Beispielen ist. Eine abstrakte Behauptung kann sich auf eine bestimmte Gesetzesbestimmung oder ein allgemeines Rechtsprinzip beziehen. Setze anschließend `nicht_abstrakt` auf `false`, wenn die Behauptung abstrakt ist, und auf `true`, wenn sie nicht abstrakt ist.
\end{lstlisting}

\subsection{Not self-advanced (\texttt{nicht\_selbst\_aufgestellt})}
\begin{lstlisting}[style=promptstyle]
Das übergreifende Projekt, in dem du genutzt wirst, ist eine Studie, in der untersucht wird, wie sich die Verwendung von Auslegungsmethoden in den Entscheidungen des Bundesverfassungsgerichts im Laufe der Geschichte entwickelt hat. Deine Aufgabe ist Teil eines Workflows, der es zum Ziel hat, einen Satz Metadaten zu erstellen, der für jede Entscheidung genau angibt, an welcher Stelle das Gericht Auslegung mittels der vier Auslegungskriterien betrieben hat.

Auslegung bedeutet, sich für eine unter mehreren möglichen Deutungen einer Gesetzesbestimmung aufgrund von Überlegungen zu entscheiden.

Aus dieser Definition lassen sich die verschiedenen Schritte ableiten, aus denen sich der Workflow zusammensetzt:

1. Deutungen identifizieren
2. Identifizieren, ob es sich bei bestimmten Sätzen überhaupt um solche handelt, die eine Begründung für die Deutung darstellen.
3. Identifizieren, welche Auslegungskriterien in dem Argument verwendet werden.

Du wirst dich ausschließlich mit dem ersten Schritt dieses Workflows befassen.

Eine Deutung ist eine Behauptung, die das Gericht selbst aufstellt und die zum Gegenstand hat, dass eine bestimmte Gesetzesbestimmung abstrakt einen bestimmten Inhalt habe.

Grundsätzlich ist hiervon auszugehen. Eine Behauptung ist jedoch dann grundsätzlich nicht selbst aufgestellt, wenn sie auf ein Präjudiz verweist (bspw. durch Zitieren einer Fundstelle wie BVerfGE 123, 456 oder den Verweis auf ständige Rechtsprechung, st. Rspr.). Wird ein Präjudiz UND andere Quellen wie zB Literaturquellen, Gesetzesbegründungen oÄ zitiert, ist die Behauptung ebenfalls selbst aufgestellt; es geht ausschließlich darum, Sätze auszunehmen, die nur Präjudizien zitieren. Der Maßstab deiner Beurteilung ist ausschließlich die Quellenlage. Unterlasse also jegliche Überlegungen zu Originalität oder Eigenleistung.

Deine Aufgabe ist es zu entscheiden, ob eine Behauptung **nicht** selbst aufgestellt ist. Der inhaltliche Maßstab ist unverändert der soeben definierte; nur die Richtung der Antwort ist umgekehrt. Das Ausgabefeld `nicht_selbst_aufgestellt` ist also genau dann `true`, wenn die Behauptung **nicht** selbst aufgestellt ist (wenn also ausschließlich Präjudizien zitiert werden), und `false`, wenn sie selbst aufgestellt ist.

Positive Beispiele (`nicht_selbst_aufgestellt: true`, die Behauptung ist also nicht selbst aufgestellt):

<example>"Deshalb muß ein Beschwerdeführer die Beseitigung des Hoheitsaktes, dessen Grundrechtswidrigkeit er geltend macht, zunächst mit den ihm durch das Gesetz zur Verfügung gestellten anderen Rechtsmitteln oder Rechtsbehelfen zu erreichen versuchen (BVerfGE 33, 192 [194]; st. Rspr.)"</example>
Hier wird ein Urteil zitiert, das anscheinend maßgeblich für die Behauptung ist, und es wird durch den Zusatz \"st. Rspr.\" (ständige Rechtsprechung) kenntlich gemacht, dass dies eine in der Rechtsprechung des Bundesverfassungsgerichts seit Langem anerkannte Position ist.

Negative Beispiele (`nicht_selbst_aufgestellt: false`, die Behauptung ist also selbst aufgestellt):

<negative_example>"Die dem Beschluß des Oberlandesgerichts Düsseldorf zugrundeliegende Auffassung, die aus dem Grundrecht des Art. 2 Abs. 1 abgeleiteten Grundsätze seien nicht von Bedeutung, verkennt die Tragweite dieser Verfassungsgarantien."</negative_example>
Hier knüpft das Gericht zwar an eine fremde Behauptung an, stellt aber seine eigene auf, indem es behauptet, dass die Auffassung des anderen Gerichts gerade nicht zuträfe.

<negative_example>**Kleine Bäckereien sind vom Nachtbackverbot ausgenommen.** Denn es ergibt sich aus Sinn und Zweck der Norm, dass sie besonders schutzbedürftig sind. Fabriken müssen ihre Mitarbeiter mit der richtigen Ausrüstung versorgen. (BVerfGE 123, 456).</negative_example>
Hier wird zwar ein Präjudiz zitiert, aber in einem anderen Satz. Daher handelt es sich bei dem Satz in Sternchen um eine selbst aufgestellte Behauptung.

Du erhältst eine Eingabe:

- `text_with_context`: ein Auszug aus einer Entscheidung des Bundesverfassungsgerichts. Der eigentlich zu beurteilende Satz ist innerhalb des Auszugs mit den Markierungen `<deutung>...</deutung>` umschlossen; die umgebenden Sätze bilden ausschließlich den Kontext.

Begründe immer deine Antwort, bevor du dich entscheidest. Beurteile kurz anhand der oben aufgestellten Maßstäbe, ob das Gericht die Behauptung selbst aufgestellt hat. Gehe dabei wie folgt vor: Nenne zunächst den Satz, der zwischen `<deutung>` und `</deutung>` steht. Liste dann ausschließlich die in diesem Satz genannten Quellen auf (Literaturquellen, Gesetzesbegründungen, Präjudizien, Gutachten uÄ). Quellen, die an anderer Stelle im Text stehen, dürfen auf keinen Fall genannt werden. Wenn keine Quellen zitiert werden, gilt die Behauptung als selbst aufgestellt. Wenn ausschließlich Präjudizien (bspw. BVerfGE 12, 345 oder 'Urteil des BVerfG vom 12.03.1993, Az. ...') zitiert werden, ist die Behauptung nicht selbst aufgestellt. In allen anderen Fällen ist die Behauptung selbst aufgestellt. Merke: (1) Entscheide ausschließlich nach Art der Quellen, nicht nach Inhalt der Behauptung und (2) Nenne ausschließlich Quellen aus dem mit `<deutung>` markierten Satz. Setze schließlich `nicht_selbst_aufgestellt` auf `true`, wenn die Behauptung nach diesem Vorgehen nicht selbst aufgestellt ist, und auf `false`, wenn sie selbst aufgestellt ist.
\end{lstlisting}

\subsection{Statutory reference (\texttt{konkretes\_gesetz})}
\begin{lstlisting}[style=promptstyle]
Das übergreifende Projekt, in dem du genutzt wirst, ist eine Studie, in der untersucht wird, wie sich die Verwendung von Auslegungsmethoden in den Entscheidungen des Bundesverfassungsgerichts im Laufe der Geschichte entwickelt hat. Deine Aufgabe ist Teil eines Workflows, der es zum Ziel hat, einen Satz Metadaten zu erstellen, der für jede Entscheidung genau angibt, an welcher Stelle das Gericht Auslegung mittels der vier Auslegungskriterien betrieben hat.

Auslegung bedeutet, sich für eine unter mehreren möglichen Deutungen einer Gesetzesbestimmung aufgrund von Überlegungen zu entscheiden.

Aus dieser Definition lassen sich die verschiedenen Schritte ableiten, aus denen sich der Workflow zusammensetzt:

1. Deutungen identifizieren
2. Identifizieren, ob es sich bei bestimmten Sätzen überhaupt um solche handelt, die eine Begründung für die Deutung darstellen.
3. Identifizieren, welche Auslegungskriterien in dem Argument verwendet werden.

Du wirst dich ausschließlich mit dem ersten Schritt dieses Workflows befassen.

Eine Deutung ist eine Behauptung, die das Gericht selbst aufstellt und die zum Gegenstand hat, dass eine bestimmte Gesetzesbestimmung abstrakt einen bestimmten Inhalt habe.

Eine Gesetzesbestimmung ist eine spezifische, abgrenzbare Norm, die durch eine genaue Bezeichnung identifiziert werden kann (bspw. § 123 BGB). Es darf sich nicht um Gewohnheitsrecht oder bloße Rechtsprinzipien/Grundsätze handeln. Wird eine Bestimmung genannt, aus der sich ein Grundsatz herleitet, zählt diese nur dann als taugliche Gesetzesbestimmung, wenn der Satz eine spezifische Behauptung über deren Inhalt enthält. Eine taugliche Gesetzesbestimmung muss nicht im Satz selbst genannt sein, sondern kann auch im Kontext stehen und implizit Gegenstand der Auslegung sein. Eine Behauptung über ein Gesetz oder eine Verordnung als Ganzes schließt eine Deutung aus.

Die Behauptung muss darüber hinaus einen bestimmten Inhalt der Gesetzesbestimmung festlegen. Das ist dann der Fall, wenn die Aussage dazu führt, dass eine Reihe von Fällen von der Gesetzesbestimmung erfasst ist und andere nicht. Nicht bestimmt ist eine Aussage hingegen, wenn sie darauf abzielt, die Gesetzesbestimmung im Ganzen als nichtig, wirksam, anwendbar oder unanwendbar zu erklären. Damit scheiden insbesondere solche Sätze aus, die nur die Vereinbarkeit einer Norm mit einer anderen zum Gegenstand haben: die Prüfung der Verfassungsmäßigkeit einer Vorschrift ebenso wie die einzelnen Schritte einer Verhältnismäßigkeitsprüfung (Geeignetheit, Erforderlichkeit, Angemessenheit) und Erwägungen zur Zweckmäßigkeit innerhalb einer solchen Prüfung. Derartige Ausführungen beziehen sich auf das Verhältnis zweier Normen zueinander und nicht auf den Inhalt der einen oder der anderen. Dass die betroffenen Vorschriften dabei genau bezeichnet werden, genügt nicht: sind die Voraussetzungen dieses Absatzes nicht erfüllt, gib eine leere Liste zurück, auch wenn im Satz eine oder mehrere Normen ausdrücklich genannt sind.

Positive Beispiele:

<example>"Das Bundeswahlgesetz ordnet in § 37 an, dass das Wahlergebnis \"nach Beendigung der Wahlhandlung\" festzustellen ist.[...] Für eine Auslegung dahingehend, dass **die \"Wahlhandlung\" am Abend der Hauptwahl beendet ist und bei der Nachwahl eine neue \"Wahlhandlung\" beginnt**, spricht aber § 46 Abs. 1 Satz 1 Nr. 2 BWG, der den Verlust des Abgeordnetenmandats für den Fall der \"Neufeststellung\" des Wahlergebnisses vorsieht."</example>
Hier wird auf § 37 BWG Bezug genommen, obwohl die Gesetzesbestimmung in der Deutung selbst nicht genannt ist.

<example>Art. 38 Abs. 1 Satz 1 GG bestimmt, dass die Abgeordneten des Deutschen Bundestages in allgemeiner, unmittelbarer, freier, gleicher und geheimer Wahl gewählt werden. Die 'Gleichheit der Wahl' erfordert dabei, dass jede Stimme den gleichen Zählwert und die gleiche Erfolgschance haben muss.</example>
Hier wird Art. 38 Abs. 1 Satz 1 GG konkret ausgelegt, indem erläutert wird, was unter der 'Gleichheit der Wahl' zu verstehen ist.

<example>Gegen die Verfassungsmäßigkeit des § 1361 Abs. 2 BGB, der durch das Gesetz über die Gleichberechtigung von Mann und Frau auf dem Gebiet des bürgerlichen Rechts vom 18. Juni 1957 (BGBl. I S. 609) neu gefaßt worden ist, bestehen weder im Hinblick auf den durch Art. 6 Abs. 1 GG geforderten Schutz von Ehe und Familie noch im Hinblick auf die Gleichberechtigung von Mann und Frau nach Art. 3 Abs. 2 GG Bedenken. Diese Unterhaltsregelung trägt beiden Verfassungsgeboten Rechnung, indem sie in den dort geregelten Fällen einen besonderen Schutz der nichterwerbstätigen Ehefrau vorsieht (vgl. den Schriftlichen Bericht des Ausschusses für Rechtswesen und Verfassungsrecht - 16. Ausschuß - zu BT-Drucks. II/3409 S. 39). <potentielle_deutung>Sie geht davon aus, daß die Frau ihre Verpflichtung zum Unterhalt der Familie in der Regel durch die Führung des Haushalts erfüllt (§ 1360 Satz 2 BGB) und häufig im Vertrauen auf die Dauerhaftigkeit der Ehe von einer außerhäuslichen Erwerbstätigkeit Abstand nimmt, um sich ausschließlich dem häuslichen Bereich der Familie zu widmen.</potentielle_deutung> Hieraus ergeben sich im Falle einer Aufhebung der ehelichen Gemeinschaft zwangsläufig erhebliche Nachteile für die erwerbswirtschaftliche Situation der Frau, und zwar auch dann, wenn sie keine Kinder zu betreuen hat (vgl. BVerfGE 17, 1 [21 f.]).</example>
Hier wird § 1361 Abs. 2 BGB konkret ausgelegt, indem erläutert wird, was unter der Gleichberechtigung von Mann und Frau zu verstehen ist. Nicht hingegen wird § 1360 Satz 2 BGB ausgelegt, obwohl er im zu analysierenden Satz zitiert wird. Der Satz trifft eine Aussage über den Inhalt von § 1361 Abs. 2 BGB und zieht zur Ausgestaltung eine Deutung aus § 1360 Satz 2 BGB heran. Dass der Kontext eine Verfassungsmäßigkeitsprüfung ist, ändert daran nichts: maßgeblich ist allein, dass der markierte Satz selbst festlegt, welche Fälle von § 1361 Abs. 2 BGB erfasst sind. Umgekehrt wird ein Satz nicht dadurch zur Deutung, dass er innerhalb einer solchen Prüfung steht.

Negative Beispiele:

<negative_example>Der Grundsatz der Gleichheit der Wahl ist im Sinne einer strengen und formalen Gleichheit zu verstehen.</negative_example>
Hier wird zwar der Grundsatz der Gleichheit der Wahl genannt; dabei handelt es sich aber um ein Rechtsprinzip, nicht um eine Gesetzesbestimmung.

<negative_example>Weder die Anordnung und die Durchführung der Nachwahl noch die Ermittlung und die Bekanntgabe des vorläufigen amtlichen Wahlergebnisses am Tage der Hauptwahl verletzten Vorschriften des Bundeswahlgesetzes oder der Bundeswahlordnung.</negative_example>
Es muss sich für eine Deutung um eine ganz bestimmte Gesetzesbestimmung handeln. Das ist hier nicht der Fall, es wird von den Vorschriften des BWG und der BWO als Ganzem gesprochen.

<negative_example>Die Nachwahl ermöglicht den Wählern in den betroffenen Wahlkreisen überhaupt erst die Teilnahme an der Wahl und verwirklicht damit den Grundsatz der Allgemeinheit der Wahl (Art. 38 Abs. 1 Satz 1 GG).</negative_example>
Die Gesetzesbestimmung wird nur als Ankerpunkt für einen Grundsatz genannt und ist damit kein tauglicher Ausgangspunkt für eine Deutung im Sinne der Definition.

<negative_example>Der Gesetzgeber muss die Wahlgrundsätze aus dem Grundgesetz beachten.</negative_example>
Hier wird auf keine spezifische Gesetzesbestimmung Bezug genommen, sondern nur allgemein auf die Wahlgrundsätze verwiesen.

<negative_example>Die Feststellung eines (vorläufigen) Ergebnisses nach der Hauptwahl und dessen Bekanntgabe durch den Bundeswahlleiter vor Durchführung der Nachwahl stellen keine Beeinträchtigung der Freiheit der Wahl dar.</negative_example>
Hier wird der allgemeine Rechtsgrundsatz der Freiheit der Wahl thematisiert, ohne dass eine spezifische Gesetzesbestimmung genannt oder eine konkrete Behauptung über deren Inhalt gemacht wird.

<negative_example>§ 901 ZPO ist infolgedessen bei feststehender Leistungsunfähigkeit des Schuldners nicht anwendbar und kann insoweit Art. 2 Abs. 2 Satz 2 GG nicht verletzen.</negative_example>
Beide Vorschriften sind genau bezeichnet, und dennoch liegt keine Deutung vor: Der Satz betrifft die Anwendbarkeit des § 901 ZPO im Ganzen und dessen Verhältnis zu Art. 2 Abs. 2 Satz 2 GG. Er legt damit nicht fest, welche Fälle von der einen oder der anderen Norm erfasst sind. Die Antwort ist eine leere Liste.

<negative_example>Die Anordnung der Haft erscheint schließlich im engeren Sinne verhältnismäßig, weil die Schwere des Eingriffs und das Gewicht der ihn rechtfertigenden Gründe in angemessenem Verhältnis zueinander stehen.</negative_example>
Es handelt sich um einen Schritt einer Verhältnismäßigkeitsprüfung. Solche Erwägungen betreffen das Verhältnis zweier Normen zueinander und haben daher keinen bestimmten Inhalt im Sinne dieser Untersuchung, auch wenn die geprüfte Vorschrift aus dem Kontext eindeutig hervorgeht.

Du erhältst eine Eingabe:

- `text_with_context_konkretes_gesetz`: ein Auszug aus einer Entscheidung des Bundesverfassungsgerichts. Der konkret zu prüfende Satz ist innerhalb des Auszugs mit den Markierungen `<deutung>...</deutung>` umschlossen; die umgebenden Sätze dienen ausschließlich als Kontext.

Begründe immer deine Antwort, bevor du dich entscheidest. Prüfe zunächst, ob die Grundaussage, die der mit `<deutung>` markierte Satz trifft, den Inhalt einer konkreten Gesetzesbestimmung ausgestaltet. Die Gesetzesbestimmung muss dabei nicht im Satz selbst genannt sein, sondern kann auch in einem anderen Satz stehen. Sobald du anfangen musst, zu rechtfertigen, dass eigentlich keine konkrete Bestimmung (sondern bspw. nur generell die Bezugnahme auf ein Gesetz als Ganzes) vorliegt, liegt keine konkrete Gesetzesbestimmung vor. Prüfe sodann, ob die Aussage einen bestimmten Inhalt festlegt, also dazu führt, dass eine Reihe von Fällen von der Bestimmung erfasst ist und andere nicht. Betrifft der Satz stattdessen die Vereinbarkeit, Wirksamkeit oder Anwendbarkeit einer Norm im Ganzen — insbesondere als Teil einer Verfassungsmäßigkeits- oder Verhältnismäßigkeitsprüfung —, so ist er nicht bestimmt und die Antwort ist eine leere Liste. Achte darauf, dass Rechtsgrundsätze als Grundaussage nur in Betracht kommen, wenn der Satz eine spezifische Behauptung über deren Inhalt enthält und sie aus einer konkreten Bestimmung hergeleitet werden. Dass mehr als eine Gesetzesbestimmung ausgelegt wird, kommt äußerst selten vor, sei entsprechend streng. Falls eine den Anforderungen entsprechende Gesetzesbestimmung vorliegt, nenne die(se) Gesetzesbestimmung(en) als Liste; falls nicht, gib eine leere Liste zurück.
\end{lstlisting}

\subsection{Argument gate (\texttt{argument})}
\begin{lstlisting}[style=promptstyle]
Das übergreifende Projekt, in dem du genutzt wirst, ist eine Studie, in der untersucht wird, wie sich die Verwendung von Auslegungsmethoden in den Entscheidungen des Bundesverfassungsgerichts im Laufe der Geschichte entwickelt hat. Deine Aufgabe ist Teil eines Workflows, der es zum Ziel hat, einen Satz Metadaten zu erstellen, der für jede Entscheidung genau angibt, an welcher Stelle das Gericht Auslegung mittels der vier Auslegungskriterien betrieben hat.

Auslegung bedeutet, sich für eine unter mehreren möglichen Deutungen einer Gesetzesbestimmung aufgrund von Überlegungen zu entscheiden.

Aus dieser Definition lassen sich die verschiedenen Schritte ableiten, aus denen sich der Workflow zusammensetzt:

1. Deutungen identifizieren
2. Identifizieren, ob es sich bei bestimmten Sätzen überhaupt um solche handelt, die eine Begründung für die Deutung darstellen.
3. Identifizieren, welche Auslegungskriterien in dem Argument verwendet werden.

Du wirst dich ausschließlich mit dem zweiten Schritt dieses Workflows befassen.

Maßstab ist hierbei das folgende:

Wird die Deutung durch das mögliche Argument begründet? Das mögliche Argument stellt nur dann eine Begründung für die Deutung dar, wenn es inhaltlich und logisch direkt mit der Deutung verknüpft ist und den gleichen Argumentationsstrang verfolgt. Sowohl Pro- als auch Contra-Argumente können eine Begründung darstellen. Sätze, die zu einem anderen Argumentationsstrang gehören oder nur thematisch ähnlich sind, sind nicht als Begründung zu betrachten. Achte besonders darauf, ob das mögliche Argument tatsächlich eine Begründung liefert, unabhängig davon, ob es vor oder nach der Deutung steht.

Du erhältst vier Eingaben:

- `deutung_sentence`: der Satz, der die Deutung enthält. Eine Deutung ist eine Behauptung, die das Gericht selbst aufstellt und die zum Gegenstand hat, dass eine bestimmte Gesetzesbestimmung abstrakt einen bestimmten Inhalt habe.
- `argument_sentence`: der zu prüfende Satz – ein potenzielles Argument, das aus der textlichen Umgebung der Deutung gegriffen wurde und sie möglicherweise begründet.
- `norm`: die Gesetzesnorm, auf die sich die Deutung bezieht.
- `text_with_context`: der umliegende Kontext, in dem Deutung und Argument auftreten.

Begründe immer deine Antwort, bevor du dich entscheidest. Erfolgt die Deutung aufgrund des `argument_sentence`?
\end{lstlisting}

\subsection{Grammatical interpretation (\texttt{wortlaut})}
\begin{lstlisting}[style=promptstyle]
Das übergreifende Projekt, in dem du genutzt wirst, ist eine Studie, in der untersucht wird, wie sich die Verwendung von Auslegungsmethoden in den Entscheidungen des Bundesverfassungsgerichts im Laufe der Geschichte entwickelt hat. Deine Aufgabe ist Teil eines Workflows, der es zum Ziel hat, einen Satz Metadaten zu erstellen, der für jede Entscheidung genau angibt, an welcher Stelle das Gericht Auslegung mittels der vier Auslegungskriterien betrieben hat.

Auslegung bedeutet, sich für eine unter mehreren möglichen Deutungen einer Gesetzesbestimmung aufgrund von Überlegungen zu entscheiden.

Aus dieser Definition lassen sich die verschiedenen Schritte ableiten, aus denen sich der Workflow zusammensetzt:

1. Deutungen identifizieren
2. Identifizieren, ob es sich bei bestimmten Sätzen überhaupt um solche handelt, die eine Begründung für die Deutung darstellen.
3. Identifizieren, welche Auslegungskriterien in dem Argument verwendet werden.

Du wirst dich ausschließlich mit dem dritten Schritt dieses Workflows befassen.

Auslegungskriterium ist hierbei das folgende:

Wortlautauslegung (=grammatische Auslegung oder auch Wortsinnauslegung) ist Auslegung nach der Bedeutung eines Ausdrucks oder einer Wortverbindung im allgemeinen oder, falls ein solcher feststellbar ist, im besonderen Sprachgebrauch der betreffenden Gesetzesbestimmung. Die Gesetzesbestimmung muss dabei nicht im Satz selbst genannt sein, sondern kann auch in einem anderen Satz stehen. Wortlautauslegung ist auch bereits dann vorhanden, wenn der Satz selbst nur ein Zitat in Anführungszeichen (bspw. „vor der Hauptwahl") enthält, die Schlussfolgerung aus dem Zitat aber in einem anderen Satz steht. Nenne dann aber auch die Schlussfolgerung. Sie liegt aber nicht vor, wenn nur das ganze Gesetz allgemein Gegenstand ist oder eine andere Gesetzesbestimmung in diesem Gesetz.

Du erhältst vier Eingaben:

- `deutung_sentence`: der Satz, der die Deutung enthält. Eine Deutung ist eine Behauptung, die das Gericht selbst aufstellt und die zum Gegenstand hat, dass eine bestimmte Gesetzesbestimmung abstrakt einen bestimmten Inhalt habe.
- `argument_sentence`: der zu prüfende Satz – ein mögliches Argument, das aus der textlichen Umgebung der Deutung gegriffen wurde und sie potenziell begründet.
- `norm`: die Gesetzesnorm, auf die sich die Deutung bezieht.
- `text_with_context`: der umliegende Kontext, in dem Deutung und Argument auftreten.

Begründe immer deine Antwort, bevor du dich entscheidest. Entspricht das `argument_sentence` der Definition und den Beispielen der Wortlautauslegung? Berücksichtige den `text_with_context` nur, um das `argument_sentence` besser zu verstehen, aber entscheide ausschließlich auf Basis des Inhalts von `argument_sentence`, ob die Auslegungsmethode angewendet wird. Wortlautauslegung liegt vor, sobald zumindest ein Teil des Wortlauts derselben Gesetzesbestimmung Gegenstand ist. Der bloße Name des ganzen Gesetzes oder eine andere Gesetzesbestimmung in diesem Gesetz als die der Deutung führen zu keiner Wortlautauslegung.
\end{lstlisting}

\subsection{Systematic interpretation (\texttt{systematik})}
\begin{lstlisting}[style=promptstyle]
Das übergreifende Projekt, in dem du genutzt wirst, ist eine Studie, in der untersucht wird, wie sich die Verwendung von Auslegungsmethoden in den Entscheidungen des Bundesverfassungsgerichts im Laufe der Geschichte entwickelt hat. Deine Aufgabe ist Teil eines Workflows, der es zum Ziel hat, einen Satz Metadaten zu erstellen, der für jede Entscheidung genau angibt, an welcher Stelle das Gericht Auslegung mittels der vier Auslegungskriterien betrieben hat.

Auslegung bedeutet, sich für eine unter mehreren möglichen Deutungen einer Gesetzesbestimmung aufgrund von Überlegungen zu entscheiden.

Aus dieser Definition lassen sich die verschiedenen Schritte ableiten, aus denen sich der Workflow zusammensetzt:

1. Deutungen identifizieren
2. Identifizieren, ob es sich bei bestimmten Sätzen überhaupt um solche handelt, die eine Begründung für die Deutung darstellen.
3. Identifizieren, welche Auslegungskriterien in dem Argument verwendet werden.

Du wirst dich ausschließlich mit dem dritten Schritt dieses Workflows befassen.

Auslegungskriterium ist hierbei das folgende:

Die systematische Auslegung erfordert: Eine spezifische, klar abgrenzbare Gesetzesbestimmung, die sich von der in der Deutung genannten unterscheidet. Bezugnahmen auf bloße Rechtsgrundsätze, Prinzipien, ganze Gesetze oder unbestimmte Vorschriften zählen nicht. Die zitierte Gesetzesbestimmung muss genutzt werden, um den Inhalt der Deutung zu erklären oder zu vertiefen; sie kann auch eine gegenteilige Position vertreten. Die Gesetzesbestimmung muss nicht im Satz selbst genannt sein, sondern kann auch in einem anderen Satz stehen.

Positive Beispiele:

<example>Der Schutzbereich des Art. 6 Abs. 1 GG umfaßt auch das Verhältnis zwischen Eltern und ihren volljährigen Kindern.
<auslegung>Schon nach der gesetzlichen Ausgestaltung erschöpft sich die Beziehung zwischen Eltern und Kindern nicht in der Erziehungsfunktion der Familie. **Die herkömmliche Regelung des Unterhaltsrechts verweist deutlich auf eine lebenslange Verpflichtung von Eltern und Kindern, einander Beistand zu leisten. Mit der Einführung von § 1618a BGB hat der Gesetzgeber als Leitbild der Eltern-Kind-Beziehung die vom Alter der Kinder unabhängige wechselseitige Pflicht zu Beistand und Rücksichtnahme statuiert.**</auslegung></example>
Indem die Textstelle auf **§ 1618a BGB** Bezug nimmt, um den Charakter des **Art. 6 Abs. 1 GG** näher zu bestimmen, berücksichtigt sie andere sachliche Bestimmungen. Diese liegen hier zwar in einer anderen Regelung als der Bezugsbestimmung, betreffen aber denselben Regelungsgegenstand und sind daher relevant für das Verständnis der Gesetzesbestimmung.

<example>**Der Begriff Ergänzungsabgabe besagt lediglich, daß diese Abgabe die Einkommen- und Körperschaftsteuer, also auf Dauer angelegte Steuern, ergänzen, d. h. in einer gewissen Akzessorietät zu ihnen stehen soll.**</example>
Um die Bedeutung des in der Vorschrift verwendeten Begriffs „Ergänzungsabgabe" näher zu bestimmen, greift der Text auf die in anderen Vorschriften des GG genannten Begriffe „Körperschaftsteuer" und „Einkommensteuer" zurück. Er berücksichtigt hier also den Kontext des Begriffs „Ergänzungsabgabe", wie er zum Verständnis des Begriffs erforderlich ist. Es handelt sich daher um systematische Auslegung.

Du erhältst vier Eingaben:

- `deutung_sentence`: der Satz, der die Deutung enthält. Eine Deutung ist eine Behauptung, die das Gericht selbst aufstellt und die zum Gegenstand hat, dass eine bestimmte Gesetzesbestimmung abstrakt einen bestimmten Inhalt habe.
- `argument_sentence`: der zu prüfende Satz – ein mögliches Argument, das aus der textlichen Umgebung der Deutung gegriffen wurde und sie potenziell begründet.
- `norm`: die Gesetzesnorm, auf die sich die Deutung bezieht.
- `text_with_context`: der umliegende Kontext, in dem Deutung und Argument auftreten.

Begründe immer deine Antwort, bevor du dich entscheidest. Vorgehensweise: Prüfe, ob das `argument_sentence` auf eine spezifische, klar abgrenzbare Gesetzesbestimmung Bezug nimmt, die sich von der in der Deutung genannten unterscheidet. Sie kann auch in den Sätzen vor dem `argument_sentence` stehen. Nenne diese Gesetzesbestimmung. Beurteile, ob diese Gesetzesbestimmung verwendet wird, um den Inhalt des `argument_sentence` direkt zu erklären oder zu vertiefen; sei hier großzügig. Systematische Auslegung liegt nur vor, wenn beide Kriterien erfüllt sind.
\end{lstlisting}

\subsection{Historical interpretation (\texttt{geschichte})}
\begin{lstlisting}[style=promptstyle]
Das übergreifende Projekt, in dem du genutzt wirst, ist eine Studie, in der untersucht wird, wie sich die Verwendung von Auslegungsmethoden in den Entscheidungen des Bundesverfassungsgerichts im Laufe der Geschichte entwickelt hat. Deine Aufgabe ist Teil eines Workflows, der es zum Ziel hat, einen Satz Metadaten zu erstellen, der für jede Entscheidung genau angibt, an welcher Stelle das Gericht Auslegung mittels der vier Auslegungskriterien betrieben hat.

Auslegung bedeutet, sich für eine unter mehreren möglichen Deutungen einer Gesetzesbestimmung aufgrund von Überlegungen zu entscheiden.

Aus dieser Definition lassen sich die verschiedenen Schritte ableiten, aus denen sich der Workflow zusammensetzt:

1. Deutungen identifizieren
2. Identifizieren, ob es sich bei bestimmten Sätzen überhaupt um solche handelt, die eine Begründung für die Deutung darstellen.
3. Identifizieren, welche Auslegungskriterien in dem Argument verwendet werden.

Du wirst dich ausschließlich mit dem dritten Schritt dieses Workflows befassen.

Auslegungskriterium ist hierbei das folgende:

Subjektiv-teleologische Auslegung (=historische Auslegung, auch Auslegung nach Entstehungsgeschichte) liegt vor bei einer Auslegung nach der Regelungsabsicht des Gesetzgebers, der die Norm geschaffen hat. Die Nennung des Wortlauts oder allgemeine Erwägungen über den Inhalt des Gesetzes stellen noch keine historische Auslegung dar. Ausführungen sind grundsätzlich dem Gericht selbst — nicht dem Gesetzgeber — zuzurechnen, wenn das Gericht dies nicht klar zum Ausdruck bringt.

Positive Beispiele:

<example>Deutung: Mit der Einführung von § 1618a BGB hat der Gesetzgeber als Leitbild der Eltern-Kind-Beziehung die vom Alter der Kinder unabhängige wechselseitige Pflicht zu Beistand und Rücksichtnahme statuiert. Mögliches Argument: Dadurch sollte insbesondere zu einer größeren Familienautonomie beigetragen und Gefährdungen der Familie als Institution entgegengewirkt werden (BTDrucks. 8/2788 S. 43).</example>
Die Textstelle nimmt Bezug auf eine Bundestagsdrucksache. In dieser kommt die Regelungsabsicht des Gesetzgebers zum Ausdruck.

Negative Beispiele:

<negative_example>Mögliches Argument: Der angefochtene Beschluß des Landgerichts Bremen beruht auch auf der Verletzung von Art. 103 Abs. 1 GG. Deutung: Es kann nicht ausgeschlossen werden, daß das Landgericht bei Beachtung von Bedeutung und Tragweite des Art. 103 Abs. 1 GG dem Beschwerdeführer Wiedereinsetzung in den vorigen Stand gewährt hätte.</negative_example>
In diesem Fall wird zwar auf die Bedeutung und Tragweite des Art. 103 Abs. 1 GG verwiesen. Daraus lässt sich aber nicht schließen, dass darin die Regelungsabsicht des Gesetzgebers zum Ausdruck kommt. Es handelt sich vielmehr um eine bloße Erwägung des Gerichts.

Du erhältst vier Eingaben:

- `deutung_sentence`: der Satz, der die Deutung enthält. Eine Deutung ist eine Behauptung, die das Gericht selbst aufstellt und die zum Gegenstand hat, dass eine bestimmte Gesetzesbestimmung abstrakt einen bestimmten Inhalt habe.
- `argument_sentence`: der zu prüfende Satz – ein mögliches Argument, das aus der textlichen Umgebung der Deutung gegriffen wurde und sie potenziell begründet.
- `norm`: die Gesetzesnorm, auf die sich die Deutung bezieht.
- `text_with_context`: der umliegende Kontext, in dem Deutung und Argument auftreten.

Begründe immer deine Antwort, bevor du dich entscheidest. Entspricht das `argument_sentence` der Definition und den Beispielen der subjektiv-teleologischen Auslegung? Liegt im `argument_sentence` nach den dargelegten Maßstäben ein subjektiv-teleologisches Argument vor, ist die Kategorie erfüllt; andernfalls nicht.
\end{lstlisting}

\subsection{Objective-teleological interpretation (\texttt{zweck})}
\begin{lstlisting}[style=promptstyle]
Das übergreifende Projekt, in dem du genutzt wirst, ist eine Studie, in der untersucht wird, wie sich die Verwendung von Auslegungsmethoden in den Entscheidungen des Bundesverfassungsgerichts im Laufe der Geschichte entwickelt hat. Deine Aufgabe ist Teil eines Workflows, der es zum Ziel hat, einen Satz Metadaten zu erstellen, der für jede Entscheidung genau angibt, an welcher Stelle das Gericht Auslegung mittels der vier Auslegungskriterien betrieben hat.

Auslegung bedeutet, sich für eine unter mehreren möglichen Deutungen einer Gesetzesbestimmung aufgrund von Überlegungen zu entscheiden.

Aus dieser Definition lassen sich die verschiedenen Schritte ableiten, aus denen sich der Workflow zusammensetzt:

1. Deutungen identifizieren
2. Identifizieren, ob es sich bei bestimmten Sätzen überhaupt um solche handelt, die eine Begründung für die Deutung darstellen.
3. Identifizieren, welche Auslegungskriterien in dem Argument verwendet werden.

Du wirst dich ausschließlich mit dem dritten Schritt dieses Workflows befassen.

Auslegungskriterium ist hierbei das folgende:

Objektiv-teleologische Auslegung bedeutet, dass die Interpretation einer Norm anhand ihres objektiven Zwecks erfolgt, wie er sich aus der heutigen Rechtsordnung und den allgemeinen Prinzipien ergibt, ohne Rückgriff auf die Absichten des historischen Gesetzgebers. Bei der objektiv-teleologischen Auslegung werden insbesondere die aktuellen Strukturen des geregelten Sachbereichs und die zugrunde liegenden rechtsethischen Prinzipien herangezogen.

Positive Beispiele:

<example>Die Regelung des § 37 BWG soll die schnelle Feststellung des Wahlergebnisses ermöglichen, um die demokratische Legitimation der gewählten Vertreter sicherzustellen. Daher ist es angemessen, das Wahlergebnis unmittelbar nach der Hauptwahl festzustellen.</example>
Hier wird der aktuelle Zweck der Norm betont, ohne historische Bezüge oder Gesetzesmaterialien zu verwenden. Es handelt sich um eine objektiv-teleologische Auslegung.

<example>Die Pflicht zur Zahlung von Steuern dient der Finanzierung staatlicher Aufgaben, die im Interesse der Allgemeinheit liegen. Deshalb ist es gerechtfertigt, dass alle Bürger entsprechend ihrer Leistungsfähigkeit zur Steuer herangezogen werden.</example>
Der objektive Zweck der Norm, nämlich die Finanzierung staatlicher Aufgaben zum Wohle der Allgemeinheit, wird hervorgehoben. Es werden keine historischen Bezüge verwendet.

Negative Beispiele:

<negative_example>Der Begriff Ergänzungsabgabe besagt lediglich, dass diese Abgabe die Einkommen- und Körperschaftsteuer ergänzt, also in einer gewissen Akzessorietät zu ihnen steht.</negative_example>
Hier wird nicht auf den Zweck oder das Ziel der Norm eingegangen, sondern lediglich auf ihre Beziehung zu anderen Steuern. Es handelt sich nicht um eine objektiv-teleologische Auslegung.

<negative_example>Die Norm wurde ursprünglich eingeführt, um nach den Ereignissen der 1950er Jahre die wirtschaftliche Stabilität zu fördern (vgl. Gesetzesbegründung von 1956). Daher ist sie so zu verstehen, dass sie auch heute noch diesem Zweck dient.</negative_example>
Obwohl hier der Zweck der Norm diskutiert wird, wird auf historische Bezüge und Gesetzesmaterialien zurückgegriffen. Es handelt sich nicht um eine rein objektiv-teleologische Auslegung gemäß der Definition.

Du erhältst vier Eingaben:

- `deutung_sentence`: der Satz, der die Deutung enthält. Eine Deutung ist eine Behauptung, die das Gericht selbst aufstellt und die zum Gegenstand hat, dass eine bestimmte Gesetzesbestimmung abstrakt einen bestimmten Inhalt habe.
- `argument_sentence`: der zu prüfende Satz – ein mögliches Argument, das aus der textlichen Umgebung der Deutung gegriffen wurde und sie potenziell begründet.
- `norm`: die Gesetzesnorm, auf die sich die Deutung bezieht.
- `text_with_context`: der umliegende Kontext, in dem Deutung und Argument auftreten.

Begründe immer deine Antwort, bevor du dich entscheidest. Entspricht das `argument_sentence` der Definition und den Beispielen des objektiv-teleologischen Auslegungskriteriums von oben? Gehe wie folgt vor: Handelt es sich um eine Auslegung nach dem Zweck oder Ziel der Norm, ohne historische Bezüge zu verwenden? Achte besonders darauf, dass das `argument_sentence` den aktuellen Zweck der Norm in der heutigen Rechtsordnung hervorhebt. Liegt ein objektiv-teleologisches Argument vor?
\end{lstlisting}

\clearpage
\section{Dataset Record Schema}
\label{app:schema}

Every record is a JSON object with three top-level keys: an integer \texttt{id},
a \texttt{row} holding the sentence(s) under classification together with their
surrounding context, and \texttt{observed} holding the gold labels for the
subtask. One stripped example from each of the two datasets follows; some
paragraph-level bookkeeping fields are elided and long text fields are truncated
with \ldots.

\paragraph{Record identity.} Every instance is identified by a small tuple of
provenance fields, all recoverable from the annotation export. A \texttt{deutungen} record (a single
candidate sentence) is identified by \texttt{(decision\_name, sentence\_id)}, where
\texttt{sentence\_id} is the \texttt{satz\_id} of that sentence. An \texttt{arguments}
record (a reading--candidate pair) is identified by \texttt{(decision\_name,
deutung\_id, argument\_start\_index)}: \texttt{deutung\_id} is the reading's stable
annotation identifier, carried through verbatim from the export and shared by every
candidate paired with that reading, and \texttt{argument\_start\_index} is the
character offset at which the candidate sentence's \texttt{satz} span begins, as recorded in the earlier
export \texttt{dataset-project-46-2026-05-14-17-12-32.json} (the \texttt{start}
field already carried on that candidate's \texttt{potential\_arguments} row, which
equals that export's \texttt{document\_structure} \texttt{satz} \texttt{start}).

\subsection{\texttt{arguments} (example from subtask \texttt{wortlaut})}
Here the \texttt{row} is a (reading, candidate) pair: \texttt{deutung\_sentence}
is the reading, \texttt{argument\_sentence} the candidate justification, and
\texttt{observed} records the gate (\texttt{argument}) and each canon label.
\begin{lstlisting}[style=promptstyle]
{
  "id": 3,
  "row": {
    "decision_name": "BVerfGE32,273",
    "deutung_id": "3tDBjYneKo",
    "deutung_sentence": "Dafür spricht schon der Wortlaut, aber auch …",
    "argument_sentence": "Gegenüber Art. 119 Abs. 3 WRV hat Art. 6 Abs. 4 GG …",
    "argument_start_index": 7391,
    "relative_position": 1,
    "text_with_context": "… <deutung>…</deutung> … <potential_argument>…</potential_argument> …",
    "norm": ["Art. 6 IV GG"]
  },
  "observed": {
    "wortlaut": 1, "systematik": 0, "geschichte": 1, "zweck": 0,
    "argument": 1,
    "reasoning": "Die Wortverbindung \"jeder Mutter\" wird herangezogen …"
  }
}
\end{lstlisting}

\subsection{\texttt{deutungen} (example from subtask \texttt{konkretes\_gesetz})}
Here the \texttt{row} is a single reading marked in \texttt{text\_with\_context};
\texttt{observed.konkretes\_gesetz} is the list of concrete provisions it
interprets (empty when none), while the reading-criteria subtasks read the
boolean fields of \texttt{observed}.
\begin{lstlisting}[style=promptstyle]
{
  "id": 2,
  "row": {
    "decision_name": "BVerfGE32,273",
    "sentence_id": 67,
    "text": "Verhält sich eine werdende Mutter nach dieser Vorschrift, …",
    "is_deutung": false,
    "text_with_context": "… <deutung>…</deutung> …",
    "norm": ["§ 5 Abs. 1 Satz 1 MuSchG"],
    "vermutlich_keine_behauptung": false
  },
  "observed": {
    "nicht_abstrakt": 0, 
    "nicht_selbst_aufgestellt": 0,
    "konkretes_gesetz": ["§ 5 Abs. 1 Satz 1 MuSchG"],
    "reasoning": ""
  }
}
\end{lstlisting}

\section{Reproducing the Scores and Benchmark Datasets}
  \label{sec:reproduction}

  To verify that the reported scores and benchmark datasets follow from the
  released artifacts and methods descriptions, we set up two tasks in the Harbor
  agent evaluation framework \citep{harbor2026}. In each, an agent independently
  implements the described procedure without access to our implementation, and a
  withheld verifier compares its output against the reference.

  For the scores, an agent working in a network-isolated container is given the 64 sets of model
  predictions, the recorded judge decisions, the test-split gold labels for all
  eight subtasks, and the parts of this paper that define the metrics
  (\cref{sec:experiments} and the table captions of \cref{app:more-models}), but
  neither our scoring code nor any reported number. The task description specifies the keys of the output table. A withheld
  verifier then recomputes every cell with our own scorer and compares it against
  the agent's output.

  The agent reproduces all 208 point estimates exactly: for each of the four
  models and two prompt conditions, precision, recall and $F_1$ on the seven
  binary subtasks, their mean, and the four statutory-reference metrics.\footnote{%
  Agreement is checked at a floating-point tolerance of $10^{-6}$; every cell
  matched exactly. The bootstrap confidence intervals are outside the check, since
  they depend on the resampling seed.} The scores in this paper are therefore
  fully determined by the protocol of \cref{sec:experiments} together with the
  published predictions, judge decisions and gold labels.

  For the benchmark datasets, the agent is given the raw annotation source files
  and the construction procedure specified in \cref{app:benchmark-construction},
  but neither our construction code nor the reference datasets. It reproduces all
  2,200 records across eight subtasks exactly, including their field values, split
  membership, and identifiers.

  The successful agent trajectories for the
  \href{https://hub.harborframework.com/jobs/35917940-c695-43ab-bdd9-17ddea56db05/trials/e8b5e216-2f0c-4cc3-b40f-d291e64c1833}{score reproduction}
  and
  \href{https://hub.harborframework.com/jobs/676116e1-8968-4245-a097-c37e6c845ed6/trials/b0f8c8ef-d598-49c1-a5dd-a4acb8d364e9}{dataset reproduction}
  are available on the Harbor Hub.
  Further discussion of this approach to reproducibility can be found in
  \href{https://www.felixringe.com/reproduce-before-publishing/}{this blog post}.


\clearpage

\section{Benchmark Construction (Methods Appendix)}
\label{app:benchmark-construction}

\noindent\emph{Note: In keeping with the reproduction approach described in \cref{sec:reproduction}, this appendix was generated entirely by an LLM.}

This appendix specifies, end to end, the deterministic transformation from the raw annotation source files to the finished benchmark. The construction is \textbf{exact}: the same source files yield the same files, down to every record, every field, and every \texttt{id}. Every pseudo-random step has a fixed generator, seed, and draw order, so "random" here never means "unpredictable". Several early definitions --- the record schema and the sentence helpers --- recur throughout.

\subsection{What the benchmark contains}
\label{app:construction-0}

The benchmark comprises eight \textbf{subtasks}, each split into three files --- \texttt{train}, \texttt{validation}, \texttt{test} --- for \textbf{24 files} in total. Each file is a JSON array of \textbf{records}, and every record has exactly three top-level keys:

\begin{lstlisting}[style=promptstyle]
{ "id": <integer>, "row": { … }, "observed": { … } }
\end{lstlisting}

The eight subtasks form two families:

\begin{itemize}
\item the \textbf{argument family} (five subtasks: \texttt{argument}, \texttt{wortlaut}, \texttt{systematik}, \texttt{geschichte}, \texttt{zweck}), where one record is one \emph{(reading, candidate-sentence)} pair;
\item the \textbf{reading family} (three subtasks: \texttt{konkretes\_\allowbreak{}gesetz}, \texttt{nicht\_\allowbreak{}abstrakt}, \texttt{nicht\_\allowbreak{}selbst\_\allowbreak{}aufgestellt}), where one record is one \emph{reading} --- a single annotated sentence.
\end{itemize}

Each subtask occupies its own directory, as \texttt{\textless{}subtask\textgreater{}/\allowbreak{}train.\allowbreak{}json}, \texttt{\textless{}subtask\textgreater{}/\allowbreak{}validation.\allowbreak{}json}, \texttt{\textless{}subtask\textgreater{}/\allowbreak{}test.\allowbreak{}json}. Only field values are significant, \texttt{id} included; the order of keys within a JSON object and the whitespace of the files are not.

\subsection{Source files}
\label{app:construction-1}

The four source files are three export forms of one Label Studio annotation project --- each preserving something the others drop, so each is consulted only for what is named below --- together with one small list.

\subsubsection{\texttt{dataset-project-46-2026-05-14-17-12-32.\allowbreak{}json} --- "structure export"}
\label{app:construction-1-1}

A JSON array of \textbf{decisions}. Each decision object carries:

\begin{itemize}
\item \texttt{decision\_\allowbreak{}name} --- the decision's identifier (a string).
\item \texttt{plain\_\allowbreak{}text} --- the decision's full text as one string. All character offsets in every file index into this string.
\item \texttt{document\_\allowbreak{}structure}, which includes (among others) a \texttt{satz} array holding the sentence segmentation. Each \texttt{satz} (sentence) object carries \texttt{satz\_\allowbreak{}id} (an integer, unique within the decision), \texttt{start} and \texttt{end} (character offsets into \texttt{plain\_\allowbreak{}text}, half-open, so the text is \texttt{plain\_\allowbreak{}text[start:end]}), \texttt{absatzID}, \texttt{ebene1nr}, \texttt{tbeg}, and \texttt{score}.
\item \texttt{potential\_\allowbreak{}deutungen} --- the array of annotated \emph{readings}. Each entry has \texttt{id} (a short opaque string, the \textbf{region id}, which is the join key across all exports), \texttt{satz\_\allowbreak{}id} (the sentence the reading sits on), \texttt{abstrakt} and \texttt{selbst\_\allowbreak{}aufgestellt} (each \texttt{true}, \texttt{false}, or \texttt{null}), and reasoning strings.
\item \texttt{potential\_\allowbreak{}arguments} --- the array of annotated \emph{argument candidates}. Each entry has \texttt{id} (a region id), \texttt{satz\_\allowbreak{}id}, \texttt{start} and \texttt{end} (offsets of the candidate sentence), \texttt{deutung\_\allowbreak{}id} (the region id of the reading the candidate is attached to), \texttt{annotation\_\allowbreak{}method} (a string, whose value \texttt{"individually"} marks a per-pair human judgment), and one object per canon (\texttt{general\_\allowbreak{}argument}, \texttt{wortlaut}, \texttt{systematik}, \texttt{geschichte}, \texttt{zweck}), each either \texttt{null} or an object \texttt{\{ "present": true|false,\allowbreak{} "reasoning": "\ldots{}" \}}.
\end{itemize}

This file is the source of all sentence text, all offsets, all context windows, and the reading/argument structure. The two remaining exports supply only the specific fields named below.

\subsubsection{\texttt{raw-labelstudio-export-project-46-2026-07-25.\allowbreak{}json} --- "raw dump"}
\label{app:construction-1-2}

The pre-merge Label Studio export: a JSON array of task objects, each with an \texttt{annotations} array, each annotation a \texttt{result} array of regions. Each region has an \texttt{id} (the same region id as above), a \texttt{from\_\allowbreak{}name} (which annotation field it is), a \texttt{type}, and a \texttt{value}. Three fields survive only here and are read only from this file:

\begin{itemize}
\item \texttt{bestimmt\_\allowbreak{}moeglicheDeutung} (type \texttt{choices}): the reading's determinate-content judgment, taken as \texttt{value.\allowbreak{}choices[0]} --- one of \texttt{"Yes"}, \texttt{"No"}, or (when empty) absent. Indexed by region id.
\item \texttt{konkretes\_\allowbreak{}gesetz\_\allowbreak{}moeglicheDeutung} (type \texttt{textarea}): the list of concrete statutory provisions the reading names. Here \texttt{value.\allowbreak{}text} is a list of strings, one per line the annotator entered; each entry is stripped of surrounding whitespace, empties are dropped, and the order is preserved. (This field must be read from this export, because the others merge the list into one unsplittable string.) Indexed by region id; a reading with no such region has the empty list.
\item \texttt{vermutlich\_\allowbreak{}keine\_\allowbreak{}behauptung\_\allowbreak{}moeglicheDeutung} (type \texttt{choices}): a boolean flag, \texttt{true} exactly when \texttt{value.\allowbreak{}choices[0] =\allowbreak{}=\allowbreak{} "Yes"}. Indexed by region id; absent means \texttt{false}.
\end{itemize}

\subsubsection{\texttt{postprocessing-export-project-46-2026-05-14-17-12-32.\allowbreak{}json} --- "label export"}
\label{app:construction-1-3}

The merged export, used only by the two reading-criteria subtasks (\texttt{nicht\_\allowbreak{}abstrakt}, \texttt{nicht\_\allowbreak{}selbst\_\allowbreak{}aufgestellt}) as the source of the human Yes/No criterion answers. Its task/annotation/result nesting matches the raw dump. Per region id it supplies:

\begin{itemize}
\item \texttt{abstrakt\_\allowbreak{}moeglicheDeutung} (type \texttt{choices}): \texttt{value.\allowbreak{}choices[0]} \ensuremath{\in} \{\texttt{"Yes"}, \texttt{"No"}\}, or absent.
\item \texttt{selbst\_\allowbreak{}aufgestellt\_\allowbreak{}moeglicheDeutung} (type \texttt{choices}): likewise.
\item \texttt{vermutlich\_\allowbreak{}keine\_\allowbreak{}behauptung\_\allowbreak{}moeglicheDeutung}: a boolean, as in \cref{app:construction-1-2}.
\item a region with \texttt{from\_\allowbreak{}name =\allowbreak{}=\allowbreak{} "label"} whose \texttt{value.\allowbreak{}labels} contains \texttt{"moeglicheDeutung"}, which marks that region id as an annotated reading.
\end{itemize}

\subsubsection{\texttt{selective\_\allowbreak{}decisions.\allowbreak{}json}}
\label{app:construction-1-4}

A JSON array of decision names: the \textbf{selectively annotated} decisions, for which not every sentence was reviewed. Every other decision in the structure export is \textbf{exhaustively annotated}. Several rules below turn on this distinction.

\subsection{Global conventions}
\label{app:construction-2}

\subsubsection{The random number generator}
\label{app:construction-2-1}

Every stochastic step uses the standard CPython \texttt{random} module generator --- the Mersenne-Twister, MT19937, i.e. \texttt{random.\allowbreak{}Random} --- with the seed fixed at \textbf{SEED = 42}. Determinism rests on three things, each fixed at its point of use: the seed, which generator instance is used where, and the order in which draws are taken. The operations, with their standard-library semantics, are \texttt{randrange(3)} (a uniform integer in \{0, 1, 2\}), \texttt{choice(seq)} (one uniformly chosen element), \texttt{shuffle(list)} (Fisher--Yates in place, the module's algorithm), and \texttt{sample(population,\allowbreak{} k)} (k distinct elements, the module's algorithm). The \texttt{id} numbering and split membership are defined by the precise MT19937 stream these produce, not by any re-implementation --- a hand-rolled Fisher--Yates or NumPy would diverge. Generators are never interleaved: wherever a step starts a fresh generator, that is a new \texttt{random.\allowbreak{}Random(42)}, independent of any other.

\subsubsection{The split order}
\label{app:construction-2-2}

Wherever the three splits are iterated --- partitioning, sampling, numbering --- they are visited in the fixed order \textbf{test, then train, then validation}. Per-split targets, written as \emph{(positives, negatives)}, are:

\begin{itemize}
\item reading family: \texttt{test =\allowbreak{} (60,\allowbreak{} 180)}, \texttt{train =\allowbreak{} (20,\allowbreak{} 60)}, \texttt{validation =\allowbreak{} (20,\allowbreak{} 60)} --- 100 positives and 300 negatives per subtask overall;
\item argument family: \texttt{test =\allowbreak{} (30,\allowbreak{} 90)}, \texttt{train =\allowbreak{} (10,\allowbreak{} 30)}, \texttt{validation =\allowbreak{} (10,\allowbreak{} 30)} --- 50 positives and 150 negatives per subtask overall.
\end{itemize}

\subsubsection{Sentence usability (\texttt{is\_\allowbreak{}evaluated})}
\label{app:construction-2-3}

The sentence segmentation emits some fragments that are not classifiable units (section headings such as "II." or "B.-I."). A sentence's text is \textbf{usable} exactly when both its length exceeds 5 characters and it contains at least one alphabetic letter from A--Z, a--z, or the German letters ä ö ü Ä Ö Ü ß. Non-usable sentences are dropped before anything else and never enter any pool.

\subsubsection{Sentence text and context windows}
\label{app:construction-2-4}

The \textbf{text} of a sentence with offsets \texttt{start}/\texttt{end} is \texttt{plain\_\allowbreak{}text[start:end]}, with every newline replaced by a single space. A \textbf{context window} around one or more target sentences, for integer counts \texttt{before} and \texttt{after}, is formed as follows. The decision's \texttt{satz} list is sorted by \texttt{satz\_\allowbreak{}id} ascending, which gives a linear sentence order and, for each \texttt{satz\_\allowbreak{}id}, an index within it. With \texttt{lo} and \texttt{hi} the smallest and largest indices among the target sentences, the window spans indices \texttt{max(0,\allowbreak{} lo - before)} through \texttt{min(last\_\allowbreak{}index,\allowbreak{} hi + after)}, inclusive. Those sentences' texts (each computed as above) are concatenated with a single space between consecutive sentences. Certain target sentences are \textbf{marked} --- wrapped with an opening and closing tag around that sentence's text before concatenation; the tags per subtask are given below. The window sizes are \texttt{DEUTUNG\_\allowbreak{}BEFORE =\allowbreak{} DEUTUNG\_\allowbreak{}AFTER =\allowbreak{} 2}, \texttt{KONKRETES\_\allowbreak{}GESETZ\_\allowbreak{}BEFORE =\allowbreak{} 10} (with its "after" still 2), and \texttt{before =\allowbreak{} after =\allowbreak{} 2} for argument windows.

\subsection{The reading family: the reading pool}
\label{app:construction-3}

Both \texttt{konkretes\_\allowbreak{}gesetz} and the two \texttt{nicht\_\allowbreak{}*} criteria begin from \emph{readings}, but along two build paths that were authored separately and read the \texttt{abstrakt} / \texttt{selbst\_\allowbreak{}aufgestellt} judgments from different files; they are described separately and do not share a pool between \cref{app:construction-3} and \cref{app:construction-5}.

\subsubsection{The \texttt{konkretes\_\allowbreak{}gesetz} pool (\texttt{build\_\allowbreak{}deutung\_\allowbreak{}entries})}
\label{app:construction-3-1}

Decisions are processed in file order and, within each, the \texttt{potential\_\allowbreak{}deutungen} in file order. A reading on sentence \texttt{satz\_\allowbreak{}id} is skipped when \texttt{satz\_\allowbreak{}id} is not among the decision's sentences, and skipped when its sentence text is not usable (\cref{app:construction-2-3}). Its \textbf{provision list} (\texttt{norms}, \cref{app:construction-1-2}) is looked up by its region \texttt{id}; denote that list \texttt{norm} (possibly empty). The reading is a \textbf{positive} exactly when \texttt{norm} is non-empty.

A non-positive reading may still become a \textbf{negative}, but only when it is a \emph{recorded} decision, which requires both of the following. First, it must have \textbf{reached the criterion}: its decision is exhaustively annotated (its \texttt{decision\_\allowbreak{}name} is not in \texttt{selective\_\allowbreak{}decisions}), and the reading's \texttt{abstrakt} is \texttt{true} and its \texttt{selbst\_\allowbreak{}aufgestellt} is \texttt{true} (both read from the structure export). Second, \texttt{bestimmt} for this region id (\cref{app:construction-1-2}) must be answered --- exactly \texttt{"Yes"} or \texttt{"No"}. When either condition fails, the reading is neither positive nor negative and is dropped. A negative's \textbf{kind} is \texttt{"kein\_\allowbreak{}bestimmter\_\allowbreak{}inhalt"} when \texttt{bestimmt =\allowbreak{}=\allowbreak{} "No"} and \texttt{"keine\_\allowbreak{}konkrete\_\allowbreak{}gesetzesbestimmung"} when \texttt{bestimmt =\allowbreak{}=\allowbreak{} "Yes"}; positives have kind \texttt{null}.

The record marks the reading's own sentence with \texttt{("\textless{}deutung\textgreater{}",\allowbreak{} "\textless{}/\allowbreak{}deutung\textgreater{}")}. Its \texttt{row} holds \texttt{decision\_\allowbreak{}name}; \texttt{ebene1nr} (the sentence's \texttt{ebene1nr}); \texttt{absatz\_\allowbreak{}id} (the sentence's \texttt{absatzID}); \texttt{sentence\_\allowbreak{}id} (\texttt{satz\_\allowbreak{}id}); \texttt{start\_\allowbreak{}index} and \texttt{end\_\allowbreak{}index} (the sentence's offsets); \texttt{tbeg}; \texttt{score} (the sentence's \texttt{score}, or \texttt{0.\allowbreak{}0} when absent); \texttt{text} (the sentence text); \texttt{should\_\allowbreak{}be\_\allowbreak{}evaluated} (\texttt{true} here, the sentence having passed \cref{app:construction-2-3}); \texttt{deutung\_\allowbreak{}id} (the region id); \texttt{is\_\allowbreak{}deutung} (always \texttt{false} for this subtask); \texttt{text\_\allowbreak{}with\_\allowbreak{}context} (the context window with \texttt{before =\allowbreak{} after =\allowbreak{} 2}, the reading sentence marked); \texttt{text\_\allowbreak{}with\_\allowbreak{}context\_\allowbreak{}konkretes\_\allowbreak{}gesetz} (the context window with \texttt{before =\allowbreak{} 10}, \texttt{after =\allowbreak{} 2}, the reading sentence marked); \texttt{norm} (the provision list); \texttt{negative\_\allowbreak{}kind} (from the kind rule above); and \texttt{vermutlich\_\allowbreak{}keine\_\allowbreak{}behauptung} (the boolean flag, \cref{app:construction-1-2}).

Its \texttt{observed} is \texttt{\{ "abstrakt": 1 if the reading's abstrakt is true else 0,\allowbreak{} "selbst\_\allowbreak{}aufgestellt": 1 if true else 0,\allowbreak{} "konkretes\_\allowbreak{}gesetz": \textless{}the provision list\textgreater{},\allowbreak{} "reasoning": \textless{}konkrete\_\allowbreak{}bezugnahme\_\allowbreak{}reasoning or ""\textgreater{} \}}. Positivity for this subtask is defined by \texttt{observed.\allowbreak{}konkretes\_\allowbreak{}gesetz} being non-empty. Each pool entry retains its \texttt{decision\_\allowbreak{}name} and, for negatives, its kind.

\subsection{The argument family: the pair pool}
\label{app:construction-4}

A single pass builds one pool of \emph{(reading, candidate)} \textbf{pairs}, from which the five argument subtasks are then cut. Decisions are processed in file order and, within each, the \texttt{potential\_\allowbreak{}arguments} in file order. A candidate with region \texttt{id}, sentence \texttt{a\_\allowbreak{}satz\_\allowbreak{}id}, offsets \texttt{start}/\texttt{end}, and \texttt{deutung\_\allowbreak{}id} is treated as follows, where its \texttt{start}/\texttt{end} are the offsets of the candidate's \textbf{annotated span} in \texttt{potential\_\allowbreak{}arguments}. The reading named by \texttt{deutung\_\allowbreak{}id} is found in the decision's \texttt{potential\_\allowbreak{}deutungen}; when there is none, the candidate is skipped. The candidate is also skipped when either the reading's sentence or the candidate's sentence is missing from the decision's sentences. A \textbf{gate on a genuine reading} then applies: the pair is kept only when the reading has \texttt{abstrakt =\allowbreak{}=\allowbreak{} true}, \texttt{selbst\_\allowbreak{}aufgestellt =\allowbreak{}=\allowbreak{} true}, and a non-empty provision list (\texttt{norms}, \cref{app:construction-1-2}, by the reading's region id).

The pair's \textbf{key} is the triple \texttt{(decision\_\allowbreak{}name,\allowbreak{} deutung\_\allowbreak{}id,\allowbreak{} candidate span start offset)}, and pairs are \textbf{deduplicated} on this key: when a pair with the same key was already kept, the new one is ignored, unless the new one is individually judged (\texttt{annotation\_\allowbreak{}method =\allowbreak{}=\allowbreak{} "individually"}) and the kept one was not, in which case the individually judged record replaces it --- so the individually judged record wins a collision, and otherwise the first seen stays. Finally, the candidate sentence text is computed, and the pair is skipped when it is not usable (\cref{app:construction-2-3}).

The record marks the reading's sentence with \texttt{("\textless{}deutung\textgreater{}",\allowbreak{} "\textless{}/\allowbreak{}deutung\textgreater{}")} and the candidate's sentence with \texttt{("\textless{}potential\_\allowbreak{}argument\textgreater{}",\allowbreak{} "\textless{}/\allowbreak{}potential\_\allowbreak{}argument\textgreater{}")}; when the two are the same sentence, that single sentence is instead wrapped \texttt{("\textless{}deutung\textgreater{}\textless{}potential\_\allowbreak{}argument\textgreater{}",\allowbreak{} "\textless{}/\allowbreak{}potential\_\allowbreak{}argument\textgreater{}\textless{}/\allowbreak{}deutung\textgreater{}")}. Its \texttt{row} holds \texttt{decision\_\allowbreak{}name}; \texttt{deutung\_\allowbreak{}id}; \texttt{ebene1nr\_\allowbreak{}deutung} and \texttt{absatz\_\allowbreak{}id\_\allowbreak{}deutung} (the reading sentence's \texttt{ebene1nr} and \texttt{absatzID}); \texttt{deutung\_\allowbreak{}sentence} (the reading sentence text); \texttt{argument\_\allowbreak{}sentence} (the text of the candidate's \textbf{sentence} \texttt{a\_\allowbreak{}satz\_\allowbreak{}id}, taken from the \texttt{satz} segmentation per \cref{app:construction-2-4}); \texttt{argument\_\allowbreak{}start\_\allowbreak{}index} and \texttt{argument\_\allowbreak{}end\_\allowbreak{}index} (the \texttt{start} and \texttt{end} of that same \texttt{satz} sentence --- its segmentation offsets, which are what these fields record even in the rare case where the candidate's annotated \texttt{start}/\texttt{end} span reaches beyond its sentence; they are not the span's offsets); \texttt{ebene1nr\_\allowbreak{}argument} and \texttt{absatz\_\allowbreak{}id\_\allowbreak{}argument} (that sentence's \texttt{ebene1nr} and \texttt{absatzID}); \texttt{relative\_\allowbreak{}position} (the candidate \texttt{satz\_\allowbreak{}id} minus the reading \texttt{satz\_\allowbreak{}id}); \texttt{text\_\allowbreak{}with\_\allowbreak{}context} (the context window over both target sentences, \texttt{before =\allowbreak{} after =\allowbreak{} 2}, both marked as above); and \texttt{norm} (the reading's provision list).

Its \texttt{observed} carries one flag per canon plus the gate flag: \texttt{wortlaut}, \texttt{systematik}, \texttt{geschichte}, and \texttt{zweck} are each \texttt{1} when that canon object on the candidate has \texttt{present =\allowbreak{}=\allowbreak{} true}, else \texttt{0}; and \texttt{argument} is \texttt{1} when \texttt{general\_\allowbreak{}argument.\allowbreak{}present =\allowbreak{}=\allowbreak{} true}, else \texttt{0}. (A \texttt{reasoning} string is added per subtask at selection time; see \cref{app:construction-6-3}.) Each pair retains its \texttt{decision\_\allowbreak{}name}, whether it is individually judged, and the raw canon objects.

\subsubsection{Cutting the five subtasks from the pair pool}
\label{app:construction-4-1}

For \texttt{argument}, the positives are the pairs with \texttt{observed.\allowbreak{}argument =\allowbreak{}=\allowbreak{} 1} and the negatives the pairs with \texttt{observed.\allowbreak{}argument =\allowbreak{}=\allowbreak{} 0} (positivity for this subtask being the \texttt{argument} flag). For each canon \texttt{c} in \{\texttt{wortlaut}, \texttt{systematik}, \texttt{geschichte}, \texttt{zweck}\}, the pool is first restricted to the \textbf{gate-positive} pairs (\texttt{observed.\allowbreak{}argument =\allowbreak{}=\allowbreak{} 1}); among those, the positives are the pairs whose canon object \texttt{c} has \texttt{present =\allowbreak{}=\allowbreak{} true}, and the negatives are the pairs whose canon object \texttt{c} exists (is not \texttt{null}) and has \texttt{present =\allowbreak{}=\allowbreak{} false}. Pairs where \texttt{c} is \texttt{null} --- the canon was never asked --- are neither, and are not eligible as negatives. (Positivity for a canon subtask is that canon's flag.)

\subsection{The reading-criteria family: \texttt{nicht\_\allowbreak{}abstrakt}, \texttt{nicht\_\allowbreak{}selbst\_\allowbreak{}aufgestellt}}
\label{app:construction-5}

These two subtasks use a \textbf{flipped polarity} --- the positive class is the rarer "No" answer to the underlying criterion. Their candidate set is drawn from the structure export but their labels from the label export (\cref{app:construction-1-3}), and they use only exhaustively annotated decisions.

\subsubsection{Candidate spans}
\label{app:construction-5-1}

Decisions are processed in file order, skipping any whose \texttt{decision\_\allowbreak{}name} is in \texttt{selective\_\allowbreak{}decisions}, and within each the \texttt{potential\_\allowbreak{}deutungen} in file order. A reading on \texttt{satz\_\allowbreak{}id} that is present among the sentences and whose sentence text is usable (\cref{app:construction-2-3}) yields a candidate span keyed by the reading's region \texttt{id}, with a \texttt{row} holding \texttt{decision\_\allowbreak{}name}, \texttt{ebene1nr}, \texttt{absatz\_\allowbreak{}id}, \texttt{sentence\_\allowbreak{}id}, \texttt{start\_\allowbreak{}index}, \texttt{end\_\allowbreak{}index}, \texttt{tbeg}, \texttt{score}, \texttt{text}, \texttt{should\_\allowbreak{}be\_\allowbreak{}evaluated} (\texttt{true}), \texttt{deutung\_\allowbreak{}id}, \texttt{is\_\allowbreak{}deutung} (which is \texttt{true} exactly when the reading's \texttt{abstrakt} is \texttt{true}, and \texttt{false} otherwise --- differing from \cref{app:construction-3-1}), \texttt{text\_\allowbreak{}with\_\allowbreak{}context} (\texttt{before =\allowbreak{} after =\allowbreak{} 2}, the reading sentence marked \texttt{("\textless{}deutung\textgreater{}",\allowbreak{} "\textless{}/\allowbreak{}deutung\textgreater{}")}), \texttt{text\_\allowbreak{}with\_\allowbreak{}context\_\allowbreak{}konkretes\_\allowbreak{}gesetz} (\texttt{before =\allowbreak{} 10}, \texttt{after =\allowbreak{} 2}), and \texttt{norm} (the provision list, \cref{app:construction-1-2}). Each span also retains the two reasoning strings from the structure export (\texttt{abstrakt\_\allowbreak{}reasoning}, \texttt{selbst\_\allowbreak{}aufgestellt\_\allowbreak{}reasoning}, each or \texttt{""}) and the provision list.

\subsubsection{Labels and polarity}
\label{app:construction-5-2}

Each criterion has a \textbf{base field} and a \textbf{flipped name}: \texttt{nicht\_\allowbreak{}abstrakt} from base field \texttt{abstrakt\_\allowbreak{}moeglicheDeutung}, and \texttt{nicht\_\allowbreak{}selbst\_\allowbreak{}aufgestellt} from base field \texttt{selbst\_\allowbreak{}aufgestellt\_\allowbreak{}moeglicheDeutung}. For each candidate span, the base field's answer is read from the label export (\cref{app:construction-1-3}) by region id. The span is an instance of the subtask only when that answer is exactly \texttt{"Yes"} or \texttt{"No"}. Under the flip, \texttt{"No"} is the subtask's positive and \texttt{"Yes"} its negative.

\subsubsection{The record's \texttt{observed} (dual encoding)}
\label{app:construction-5-3}

Each reading-criteria record carries both criteria's encodings --- regardless of which subtask file it lands in --- together with the provision list. Its \texttt{observed} has \texttt{konkretes\_\allowbreak{}gesetz} equal to the span's provision list. For each of the two criteria (with its base field, its base key \texttt{abstrakt} or \texttt{selbst\_\allowbreak{}aufgestellt}, and its flipped name), the base field's answer for the span is read from the label export: when it is \texttt{"Yes"} or \texttt{"No"}, \texttt{observed[base\_\allowbreak{}key]} is \texttt{1 if "Yes" else 0} and \texttt{observed[flipped\_\allowbreak{}name]} is \texttt{1 \ensuremath{-} that}; when it is neither, both \texttt{observed[base\_\allowbreak{}key]} and \texttt{observed[flipped\_\allowbreak{}name]} are \texttt{null}. The \texttt{reasoning} field concatenates, for each criterion that was answered and whose corresponding reasoning string (\cref{app:construction-5-1}) is non-empty, the fragment \texttt{"[\textless{}base\_\allowbreak{}key\textgreater{}] \textless{}reasoning\textgreater{}"}, joined by single spaces, with the \texttt{abstrakt} criterion first and then \texttt{selbst\_\allowbreak{}aufgestellt}. The \texttt{row} is the span's row (\cref{app:construction-5-1}) with one field added: \texttt{vermutlich\_\allowbreak{}keine\_\allowbreak{}behauptung}, the boolean flag for the span read from the label export (absent means \texttt{false}).

\subsection{Selecting, splitting, ordering, and numbering}
\label{app:construction-6}

The steps above define, per subtask, a \textbf{positive pool} and a \textbf{negative pool} of records, each tagged with its \texttt{decision\_\allowbreak{}name}. The remaining steps choose which records enter the benchmark, assign them to splits, order them, and number them. The two families use different split-search and sampling procedures, each described below.

Two rules are common to all subtasks. First, \textbf{splits are decision-disjoint}: every record from a given decision goes to exactly one split, so splitting is an assignment of whole decisions to \texttt{test}, \texttt{train}, or \texttt{validation}. Second, a \textbf{prompt-example hold-out} applies: a few decisions are quoted verbatim in a subtask's worked-example prompt and therefore never appear in that subtask's \texttt{test} split. The forbidden-from-test decisions are, per subtask, \texttt{BVerfGE61,\allowbreak{}126} and \texttt{cs20090421\_\allowbreak{}2bvc000206} for \texttt{konkretes\_\allowbreak{}gesetz}; \texttt{BVerfGE57,\allowbreak{}170} for \texttt{systematik}; \texttt{BVerfGE57,\allowbreak{}170} for \texttt{geschichte}; \texttt{BVerfGE57,\allowbreak{}170}, \texttt{BVerfGE61,\allowbreak{}126}, and \texttt{BVerfGE62,\allowbreak{}338} for \texttt{nicht\_\allowbreak{}abstrakt}; \texttt{BVerfGE57,\allowbreak{}170} for \texttt{nicht\_\allowbreak{}selbst\_\allowbreak{}aufgestellt}; and none for the remaining subtasks (\texttt{argument}, \texttt{wortlaut}, \texttt{zweck}).

\subsubsection{\texttt{konkretes\_\allowbreak{}gesetz}: split, then sample per negative kind}
\label{app:construction-6-1}

\textbf{Targets.} \texttt{konkretes\_\allowbreak{}gesetz} is a reading-family subtask (\cref{app:construction-0}) and therefore uses the reading-family per-split targets of \cref{app:construction-2-2}: \texttt{test =\allowbreak{} (60,\allowbreak{} 180)}, \texttt{train =\allowbreak{} (20,\allowbreak{} 60)}, \texttt{validation =\allowbreak{} (20,\allowbreak{} 60)} --- 100 positives and 300 negatives in all. It does not take the argument-family targets \texttt{(30,\allowbreak{} 90)/\allowbreak{}(10,\allowbreak{} 30)/\allowbreak{}(10,\allowbreak{} 30)}, which belong only to \texttt{argument} and the four canons even though \cref{app:construction-6-3} reuses this subtask's split-search and sampler.

\textbf{Kind targets.} Because the two negative kinds are not interchangeable, each split holds them at the pool's overall proportion. With \texttt{total\_\allowbreak{}neg} the size of the whole negative pool and \texttt{rare} the count of kind \texttt{keine\_\allowbreak{}konkrete\_\allowbreak{}gesetzesbestimmung} within it, a split with negative target \texttt{tn} has a rare-kind target of \texttt{round(tn * rare /\allowbreak{} total\_\allowbreak{}neg)} and a \texttt{kein\_\allowbreak{}bestimmter\_\allowbreak{}inhalt} target of \texttt{tn} minus that; \texttt{round} is banker's rounding, i.e. Python's built-in \texttt{round}.

\textbf{Split search (the "random-restart" search).} The search ranges only over the \textbf{pool decisions}: the distinct \texttt{decision\_\allowbreak{}name}s that own at least one record in this subtask's pool (its positives or its negatives). A decision with no record for this subtask is not a pool decision and never enters the search; in particular it spends no draw. Each pool decision has counts of positives, negatives, and negatives of each kind. A fresh \texttt{random.\allowbreak{}Random(42)} is started, and the following search runs for a fixed 300000 iterations, keeping the best assignment found. In each iteration:

\begin{enumerate}
\item with \texttt{names} the pool decisions sorted ascending as strings, one \texttt{randrange(3)} is drawn per name in that order, giving each decision a split index (0 = test, 1 = train, 2 = validation) --- exactly one draw is spent per name, so \texttt{names} is precisely the pool decisions, no dataset-wide decisions that lack a record here, or the whole random stream shifts;
\item each forbidden-from-test decision (taken in sorted order) currently assigned to test is reassigned with \texttt{choice([1,\allowbreak{} 2])};
\item the per-split totals are tallied --- positives, negatives, and per-kind negatives;
\item the assignment is feasible exactly when, for every split, positives \ensuremath{\geq} its positive target, negatives \ensuremath{\geq} its negative target, and each negative kind's count \ensuremath{\geq} that split's kind target;
\item its score is the minimum, over all splits, of \texttt{(positives \ensuremath{-} positive target)} and \texttt{(negatives \ensuremath{-} negative target)} --- the minimum slack.
\end{enumerate}

Among feasible assignments, the one with the largest score is kept, ties going to the one found earlier in the iteration. The result maps each split to its sorted list of decision names. (The search is robust to the order in which forbidden decisions are visited; the sorted order fixes the procedure concretely.)

\textbf{Sampling.} A single fresh \texttt{random.\allowbreak{}Random(42)} is then started and threaded through the splits in the order test, train, validation. For each split: the split's positive target is taken, via the deterministic sampler below, from the positive records whose decision is in the split; then, for each negative kind in ascending alphabetical order of the kind name (\texttt{kein\_\allowbreak{}bestimmter\_\allowbreak{}inhalt}, then \texttt{keine\_\allowbreak{}konkrete\_\allowbreak{}gesetzesbestimmung}), that kind's target is taken from the negative records of that kind whose decision is in the split, using the same sampler and the same generator; and the positives and the two kinds' negatives are concatenated and then shuffled with the same generator, that shuffled order being the file order.

\textbf{The deterministic sampler} \texttt{take(pool,\allowbreak{} n)}, using the current generator, sorts \texttt{pool} by the pair \texttt{(record.\allowbreak{}row.\allowbreak{}decision\_\allowbreak{}name,\allowbreak{} canonical JSON of record.\allowbreak{}row)} --- where "canonical JSON" is the JSON serialization of the \texttt{row} object with keys sorted, equivalent to Python's \texttt{json.\allowbreak{}dumps(row,\allowbreak{} sort\_\allowbreak{}keys=\allowbreak{}True)} --- then shuffles the sorted list with the generator and takes the first \texttt{n}. Because the single generator is threaded through test, then train, then validation, the sequence of \texttt{shuffle} calls per split is positives, kind-1 negatives, kind-2 negatives, whole-split shuffle, repeated for the three splits in order.

\subsubsection{\texttt{konkretes\_\allowbreak{}gesetz}: numbering}
\label{app:construction-6-2}

Within each split file, records are numbered \texttt{id =\allowbreak{} 1,\allowbreak{} 2,\allowbreak{} 3,\allowbreak{} \ldots{}} in the file order fixed by the final per-split shuffle above.

\subsubsection{The argument family (\texttt{argument} and the four canons): split and sample}
\label{app:construction-6-3}

Each of the five subtasks is split and sampled independently, by the same procedure as \texttt{konkretes\_\allowbreak{}gesetz} (the same random-restart search and the same sampler), minus the kind machinery.

\textbf{Split search.} For each pool decision (as in \cref{app:construction-6-1}: only decisions that own a positive or negative record for this subtask, never the full dataset), its positives and negatives for this subtask are counted, and for \texttt{argument} its individually judged negatives as well. The random-restart search of \cref{app:construction-6-1} runs identically --- a fresh \texttt{random.\allowbreak{}Random(42)}, 300000 iterations, \texttt{names} the pool decisions sorted ascending, one \texttt{randrange(3)} per name, the same forbidden-from-test handling, maximizing the minimum slack, earliest tie kept --- with the argument-family targets, and with one addition for the \texttt{argument} subtask only: an assignment is feasible only when, in every split, the number of individually judged negatives assigned to that split does not exceed the split's negative target. (This forces every individually judged negative to fit, so all of them enter the benchmark.) The canon subtasks add no such constraint.

\textbf{Sampling.} A single fresh \texttt{random.\allowbreak{}Random(42)} is started, and for each split in the order test, train, validation: the split's positive target is taken, via the deterministic sampler, from this subtask's positive records whose decision is in the split; the split's negative target is taken from this subtask's negative records whose decision is in the split --- for \texttt{argument} via the priority sampler below, for the four canons via the plain sampler; and the positives and negatives are concatenated and shuffled with the same generator, giving the file order.

\textbf{The priority sampler} (for \texttt{argument} negatives) sorts the pool by \texttt{(decision\_\allowbreak{}name,\allowbreak{} canonical JSON of row)} as before, separates it into the individually judged records and the rest, shuffles the individually judged group with the generator, shuffles the rest with the generator, places the individually judged group first followed by the rest, and takes the first \texttt{n} --- so the two \texttt{shuffle} calls occur in that order, then the take.

\textbf{Reasoning.} When a pair is selected into a subtask, \texttt{observed.\allowbreak{}reasoning} is added: for \texttt{argument} it is \texttt{general\_\allowbreak{}argument.\allowbreak{}reasoning} (or \texttt{""}), and for a canon \texttt{c} it is that canon object's \texttt{reasoning} (or \texttt{""}).

\textbf{Numbering.} Within each split file, records are numbered \texttt{id =\allowbreak{} 1,\allowbreak{} 2,\allowbreak{} 3,\allowbreak{} \ldots{}} in the file order fixed by the final per-split shuffle.

\subsubsection{The reading-criteria family (\texttt{nicht\_\allowbreak{}abstrakt}, \texttt{nicht\_\allowbreak{}selbst\_\allowbreak{}aufgestellt}): a different split search, sampler, and numbering}
\label{app:construction-6-4}

These two subtasks use an exhaustive decision-disjoint split search (not the random restart), a different sampler, and --- importantly --- a \textbf{global} \texttt{id} numbering, each subtask handled independently.

\textbf{Split search (exhaustive, deterministic).} For each decision, \texttt{(p,\allowbreak{} g)} are its counts of this subtask's positives and negatives. The decisions are ordered by \texttt{(\ensuremath{-}p,\allowbreak{} decision\_\allowbreak{}name)} --- most positives first, ties by name ascending. All assignments of decisions to the three splits are searched (a decision in the forbidden-from-test set may go only to train or validation), pruning a branch as soon as the positives still available among the not-yet-assigned decisions can no longer cover the remaining positive deficit across splits. Among all feasible assignments (every split meeting both its positive and negative target), the one chosen maximizes, lexicographically, the pair \emph{(minimum positive slack across splits, minimum negative slack across splits)}, ties going to the one reached earlier in this ordered enumeration. The targets are read in split order test, train, validation.

\textbf{Sampling.} A single fresh \texttt{random.\allowbreak{}Random(42)} is started, and for each split in the order test, train, validation: the split's positive pool is all this subtask's positive records whose decision is in the split, sorted by the record's region id (\texttt{deutung\_\allowbreak{}id}) ascending, and likewise the negative pool sorted by region id; the split's positive target is drawn from the sorted positive pool with \texttt{sample}, then its negative target from the sorted negative pool with \texttt{sample} (the same generator, positives first); and the sampled positives and negatives are concatenated and shuffled with the same generator, giving the split's internal order.

\textbf{Global numbering} (the distinctive part). After all three splits are sampled, shuffled, and turned into records, the three splits' record lists are concatenated in the order test, train, validation into one list; that combined list is shuffled with the same generator; and the combined list is numbered \texttt{id =\allowbreak{} 1,\allowbreak{} 2,\allowbreak{} \ldots{}} up to the total (each of these two subtasks has 400 records: 100 positives and 300 negatives). Each record keeps the number it receives here as its \texttt{id} but stays physically in its own split file, in the per-split order from the previous step --- so within a reading-criteria file the \texttt{id}s are not \texttt{1.\allowbreak{}.\allowbreak{}N} in order but the scattered global numbers. (Only these two subtasks number globally; the argument family and \texttt{konkretes\_\allowbreak{}gesetz} number \texttt{1.\allowbreak{}.\allowbreak{}N} per file.) Concretely, the generator's call sequence for one reading-criteria subtask is \texttt{sample}(test positives), \texttt{sample}(test negatives), \texttt{shuffle}(test), \texttt{sample}(train positives), \texttt{sample}(train negatives), \texttt{shuffle}(train), \texttt{sample}(validation positives), \texttt{sample}(validation negatives), \texttt{shuffle}(validation), \texttt{shuffle}(combined).

\subsection{The finished benchmark}
\label{app:construction-7}

The benchmark is the 24 files \texttt{\textless{}subtask\textgreater{}/\allowbreak{}\{train,\allowbreak{}validation,\allowbreak{}test\}.\allowbreak{}json}, each a JSON array of the \texttt{\{ id,\allowbreak{} row,\allowbreak{} observed \}} records built above, in the file order and with the \texttt{id} numbering fixed above. The eight subtask directory names are exactly \texttt{argument}, \texttt{wortlaut}, \texttt{systematik}, \texttt{geschichte}, \texttt{zweck}, \texttt{konkretes\_\allowbreak{}gesetz}, \texttt{nicht\_\allowbreak{}abstrakt}, and \texttt{nicht\_\allowbreak{}selbst\_\allowbreak{}aufgestellt}.

\end{document}